\documentclass{article} 
\usepackage{iclr2024_conference,times}

\usepackage{amsmath,amsfonts,bm}

\def\Figref#1{Figure~\ref{#1}}

\def\Secref#1{Section~\ref{#1}}

\def\eqref#1{equation~\ref{#1}}
\def\Eqref#1{Equation~\ref{#1}}

\def\1{\bm{1}}

\DeclareMathAlphabet{\mathsfit}{\encodingdefault}{\sfdefault}{m}{sl}
\SetMathAlphabet{\mathsfit}{bold}{\encodingdefault}{\sfdefault}{bx}{n}

\usepackage{hyperref}
\usepackage{url}
\usepackage{xspace}
\usepackage{dsfont}
\usepackage{graphicx}
\usepackage{booktabs}
\usepackage{multirow}
\usepackage{hhline}
\usepackage{colortbl}
\usepackage{wrapfig}

\newcommand{\tableref}[1]{Table~\ref{#1}}

\makeatletter
\DeclareRobustCommand\onedot{\futurelet\@let@token\@onedot}
\def\@onedot{\ifx\@let@token.\else.\null\fi\xspace}
\def\eg{\emph{e.g}\onedot} 
\def\ie{\emph{i.e}\onedot}

\def\@fnsymbol#1{%
   \ifcase#1\or
   \TextOrMath \textdagger \dagger\or
   \TextOrMath \textsection  \mathsection\or
   \TextOrMath \textparagraph \mathparagraph\or
   \TextOrMath \textbardbl \|\or
   \TextOrMath {\textdagger\textdagger}{\dagger\dagger}\or
   \TextOrMath {\textdaggerdbl\textdaggerdbl}{\ddagger\ddagger}\else
   \@ctrerr \fi
}
\makeatother

\title{LRR: Language-Driven Resamplable\\Continuous Representation against\\Adversarial Tracking Attacks}

\author{Jianlang Chen$^{1}$ \quad Xuhong Ren$^{2}$ \quad
Qing Guo$^{3}$\thanks{Qing Guo is the corresponding author (\href{mailto:tsingqguo@ieee.org}{tsingqguo@ieee.org})} \quad Felix Juefei-Xu$^{4}$\thanks{Work done prior to joining Meta.} \quad Di Lin$^{5}$ \\ \textbf{Wei Feng$^{5}$ \quad Lei Ma$^{6,7}$ \quad Jianjun Zhao$^{1}$} \\
$^1$ Kyushu University, Japan \quad $^2$ Tianjin University of Technology, China \\
$^3$ CFAR and IHPC, Agency for Science, Technology and Research (A*STAR), Singapore \\
$^4$ GenAI, Meta, USA \quad $^5$ Tianjin University, China \quad  $^6$ The University of Tokyo, Japan \\
$^7$ University of Alberta, Canada
}

\iclrfinalcopy 
\begin{document}

\maketitle

\begin{abstract}
Visual object tracking plays a critical role in visual-based autonomous systems, as it aims to estimate the position and size of the object of interest within a live video. Despite significant progress made in this field, state-of-the-art (SOTA) trackers often fail when faced with adversarial perturbations in the incoming frames. This can lead to significant robustness and security issues when these trackers are deployed in the real world.
To achieve high accuracy on both clean and adversarial data, we propose building a spatial-temporal implicit representation using the semantic text guidance of the object of interest extracted from the language-image model (\ie, CLIP). This novel representation enables us to reconstruct incoming frames to maintain semantics and appearance consistent with the object of interest and its clean counterparts.
As a result, our proposed method successfully defends against different SOTA adversarial tracking attacks while maintaining high accuracy on clean data. In particular, our method significantly increases tracking accuracy under adversarial attacks with around 90\% relative improvement on UAV123, which is close to the accuracy on clean data. 
We have built a benchmark and released our code in \url{https://github.com/tsingqguo/robustOT}.
\end{abstract}

\section{Introduction}
\label{intro}

Visual object tracking is a crucial technique in the field of vision intelligence, predicting the position and size of targeted objects in real-time video. It has found applications in various autonomous systems, including self-driving cars, unmanned aircraft, and robotics
Over the years, significant advancements have been made in visual object tracking. State-of-the-art tracking methods now achieve high accuracy on challenging datasets by utilizing fully trained deep neural networks (DNNs).
However, similar to the vulnerability of DNNs in image classification \citep{goodfellow2014explaining,carlini2017towards,guo2020watch}, deep tracking methods also face similar challenges \citep{wiyatno2019physical,jia2021iou, yan2020cooling,liang2020efficient,yin2022dimba}. Adversarial attacks can exploit this vulnerability by adding imperceptible perturbations to incoming frames, leading to incorrect predictions of the object's position by the deployed trackers. Such attacks pose security risks when deep trackers are integrated into automatic systems.
These attacks could cause security issues when we embed deep trackers into the automatic systems.
Therefore, it is crucial to enhance the robustness of deep trackers against adversarial tracking attacks.

There are two primary approaches to enhancing adversarial robustness in the context of image classification tasks. These include adversarial training \citep{kurakin2016adversarial,tramer2017ensemble,rebuffi2021fixing} and image preprocessing \citep{yuan2020ensemble,nie2022diffusion,ho2022disco}. However, directly applying these methods to defend against adversarial tracking attacks is not straightforward. 
Adversarial training involves retraining deep models using a min-max optimization strategy, where the DNNs are exposed to more adversarial examples during the training process. However, this approach has certain limitations, such as a potential sacrifice in accuracy on clean data and increased time costs for training.
Existing image preprocessing methods neglect the video sequence's temporal and the object template's semantic information, inadequately addressing the challenges of adversarial tracking attacks.

\begin{figure}[t]
\centering
\includegraphics[width=0.8\linewidth]{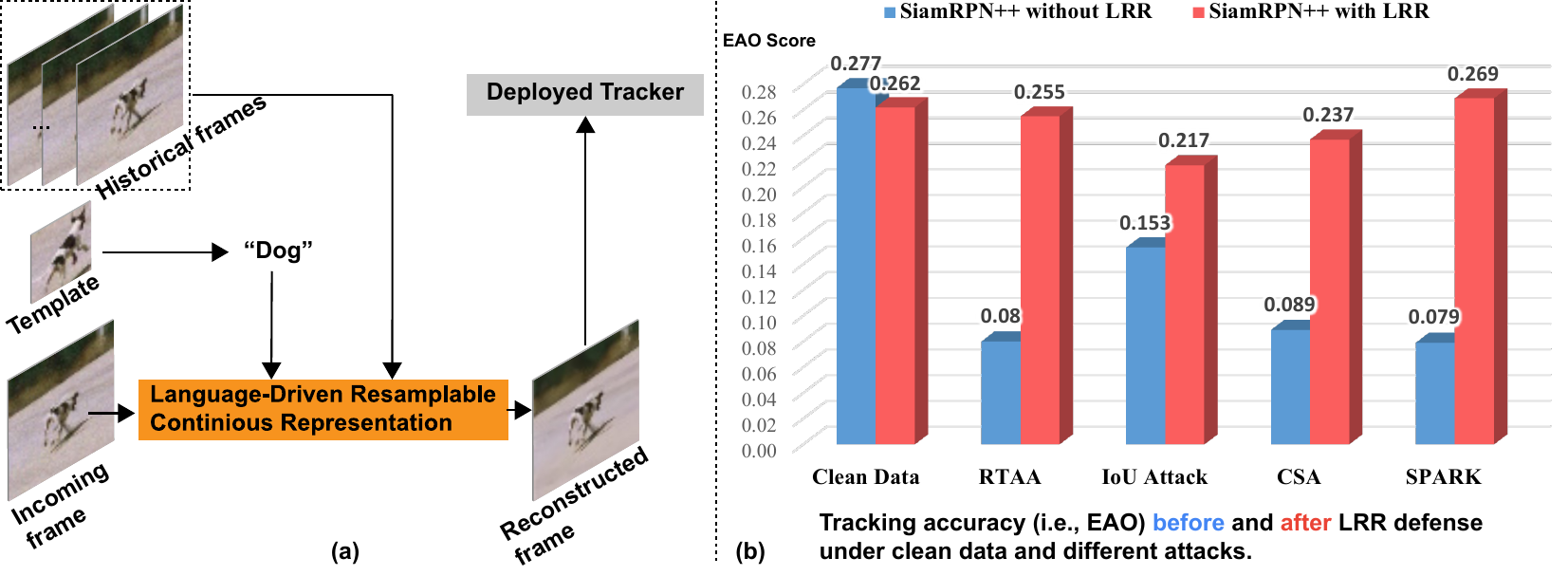}
\vspace{-10pt}
   \caption{ (a) shows the main idea of this work: we propose the language-driven resamplable continuous representation (LRR) that takes the template's text term and historical frames as inputs to reconstruct the incoming frame. (b) shows the results on VOT2019 \citep{vot2019} with and without LRR under clean data and different attacks.}
\label{fig:motivation}
\vspace{-10pt}
\end{figure}

In this study, our focus is on a preprocessing-based solution to defend against tracking attacks. Specifically, we reconstruct the incoming frames and provide them to the deployed trackers to enhance adversarial robustness (See \Figref{fig:motivation} (a)).
We argue that an effective preprocessing defense against tracking attacks should fulfill two criteria: (1) it should fully leverage the spatial and temporal contexts, which offer complementary appearance information, and (2) it should maintain semantic consistency with the object of interest as indicated by the initial frame, known as the object template.
To achieve these objectives, we propose an approach based on the implicit representation \citep{chen2021learning}, which effectively models the appearance of pixels based on their neighboring pixels. 
While existing implicit representation methods have shown promising results in image restoration, we propose a novel \textit{language-driven resamplable continuous representation (LRR)} consisting of two key modules.
First, we introduce the spatial-temporal implicit representation (STIR), enabling the reconstruction of any pixel at continuous spatial and temporal coordinates. This capability allows for the effective removal of adversarial perturbations and the achievement of appearance consistency with clean frames.
Second, we propose a language-driven resample network (LResampleNet) that leverages the STIR. This network generates a new frame by feeding resampled continuous coordinates to the STIR, guided by the text from the object template. By aligning the resampled frame with the semantic information provided by the object template, we achieve semantic consistency.
We conducted extensive experiments on three public datasets, demonstrating that our method significantly enhances the adversarial robustness of object trackers against four state-of-the-art adversarial attacks. Moreover, our approach maintains high accuracy on clean data, with the adversarial accuracy even matching or surpassing the clean accuracy. For instance, in the VOT 2019 results shown in \Figref{fig:motivation} (b), SiamRPN++ with LRR achieves an EAO of 0.283 under the SPARK attack, outperforming the 0.079 EAO achieved by SiamRPN++ without LRR and even surpassing the results on clean data.

%

\section{Background and Related Works}
\label{sec:relatedwork}

{\bf Visual object tracking.} %
Siamese trackers have become the current trend in visual object tracking tasks since they strike a great balance between tracking accuracy and efficiency \citep{li2018high,zhang2019deeper,fu2021onboard,cao2021siamapn++}. 
The SiamRPN \citep{li2018high} algorithm approaches VOT as a one-shot detection problem and was the first to introduce a region proposal network (RPN \citep{ren2015faster}) into the tracking arena. 
By incorporating RPN, SiamRPN mitigates the need for heavy multi-scale correlation operations, resulting in high-speed and accurate tracking performance. %
SiamRPN+ \citep{zhang2019deeper} and SiamRPN++ \citep{li2019siamrpn++} propose the incorporation of a cropping residual unit and a spatial-aware sampling strategy, enabling the Siamese RPN framework to benefit from modern backbones and significantly enhance the performance of the Siamese tracker. 
In this work, we evaluate the effectiveness of our defense mechanism on two trackers from the SiamRPN++ family that are popular within adversarial research. In recent years, transformer-based trackers \citep{ye2022joint,lin2022swintrack, cui2022mixformer, mayer2022transforming} have demonstrated remarkable tracking accuracy. Our initial results indicate that our method remains effective for transformer-based trackers.

{\bf Adversarial tracking attacks.}
\label{subsec:relatedwork2}
In recent years, the broad applications of visual object tracking have prompted a wide range of studies on the robustness of visual object trackers \citep{wiyatno2019physical,guo2019selective}. 
AD2Atk \citep{fu2022ad} focuses on generating adversarial examples during the resampling of the search path image.
EfficientAdv \citep{liang2020efficient} presents an end-to-end network that employs a novel drift loss in conjunction with the embedded feature loss to attack the tracker. 
DIMBA \citep{yin2022dimba} proposes a black-box attack that uses reinforcement learning to localize crucial frames accurately.
CSA \citep{yan2020cooling} employs a well-crafted cooling-shrinking loss to train an efficient adversarial perturbation generator. 
RTAA \citep{jia2020robust} conducts a frame-by-frame attack, introducing temporal perturbation into the original video sequences and significantly reducing the tracking performance. 
SPARK \citep{guo2020spark} is designed to attack online trackers by imposing a $L_p$ constraint on perturbations while calculating them incrementally based on previous attacks. 
IoU \citep{jia2021iou} creates perturbations by utilizing temporally correlated information and incrementally adding noise from the initial frame to subsequent frames.

These advanced attackers exploit the unique characteristics of VOT, thereby making defense methods, originally adapted from the image classification domain, difficult to apply effectively. 
In response to this, our approach seeks to use spatial-temporal representation to leverage the information concealed within inter-frame relationships.

{\bf Adversarial robustness enhancement}
Approaches for enhancing robustness typically fall into two main categories: adversarial training and input preprocessing during testing time. 
The adversarial training approach introduces adversarial perturbations during training \citep{goodfellow2014explaining,kurakin2016adversarial,madry2017towards,tramer2017ensemble,athalye2018obfuscated,rebuffi2021fixing}, which are usually computationally expensive.
The input preprocessing methods \citep{yuan2020ensemble,nie2022diffusion,ho2022disco} are employed to remove the adversarial perturbations, and thus enhance the robustness. However, these methods are mainly designed for image classification tasks and cannot be used to defend against adversarial tracking attacks directly.
For example, DiffPure \citep{nie2022diffusion} utilizes diffusion models for adversarial purification. 
While it exhibits promising results in image classification tasks, its intensive computational demands make it infeasible for video tasks. 
The purification process for a single image of $256\times256$ pixels requires approximately 26 seconds, which equates to a processing speed of 0.04 fps for video frame processing. We provide an empirical study in \ref{app:diffpure} by using the DiffPure for tracking defense.
Unlike previous enhancement approaches, our method leverages historical information from the object template to build a novel defense pipeline against video-specific adversarial attacks.

{\bf Implicit representation.} 
Implicit representation has been extensively employed in the modeling of 3D object shapes and structures, where a 3D object is typically represented by a multilayer perceptron (MLP) that maps coordinates to signals. 
Inspired by its success in 3D tasks, recent studies have proposed the application of implicit representation in 2D tasks. 
\citep{chen2021learning} proposed the Local Implicit Image Function (LIIF), which generates a continuous representation for super-resolution.
\cite{lee2022cvpr} improves LIIF by adding high-frequency information in Fourier space.
\citep{ho2022disco} emerged with an adversarial defense method that eliminates adversarial perturbations by utilizing local implicit functions. 
Both DISCO and LIIF perform their tasks based on local implicit image representation. 
Contrastingly, our work proposes a novel approach that extends local implicit representation into spatial-temporal implicit representation.

\section{Language-Driven Resamplable Continuous Representation}
\label{sec:method}

\subsection{Overview}
\label{subsec:method-overview}

Given a live video, an object tracker aims to predict the position and size of the object of interest, which is indicated by an object template $\mathbf{T}$ cropped from the first frame. 
Adversarial tracking attacks usually inject adversarial perturbations into incoming frames, leading to incorrect tracking results.
In this section, we propose the \textit{language-driven resamplable continuous representation (LRR)} against adversarial tracking attacks. 
The intuitive idea is that we try to reconstruct an incoming frame to remove the penitential adversarial perturbations while maintaining its semantic consistency to  the object template indicated in the first frame.
Given an incoming frame $\mathbf{I}_t$ that may be corrupted by adversarial perturbation, we try to reconstruct it and get a new counterpart $\hat{\mathbf{I}}_t$.
The objective contains two components: the first one is to remove the adversarial perturbations and encourage the reconstructed frame to have the same appearance as its clean counterpart.
The second component is to make the semantic information of the reconstructed frame and the object template be consistent. 

We have the following challenges when addressing the two objectives: \textit{First}, as we are handling a live video, the historical frames should provide complementary spatial and temporal information, that is, a perturbed pixel usually has a similar appearance to its spatial and temporal neighboring pixels that can be used to reconstruct the perturbed pixels. \textit{The key problem is how to build a bridge between the spatial \& temporal axes and pixel appearances,} which should have a high generalization to adapt to different pixel intensities or colors. \textit{Second}, in terms of semantic consistency, a straightforward solution is to extract the deep features (\eg, VGG features) of the incoming frame and the object template, respectively, and then encourage the two features to be similar. However, such a solution could only approach deep feature consistency instead of semantic consistency. There are two reasons preventing this solution: (1) the deep features are not exactly aligned with the semantic space. (2) the deep features themselves are still vulnerable to adversarial perturbations.

\begin{figure}[t]
\centering
\includegraphics[width=0.9\linewidth]{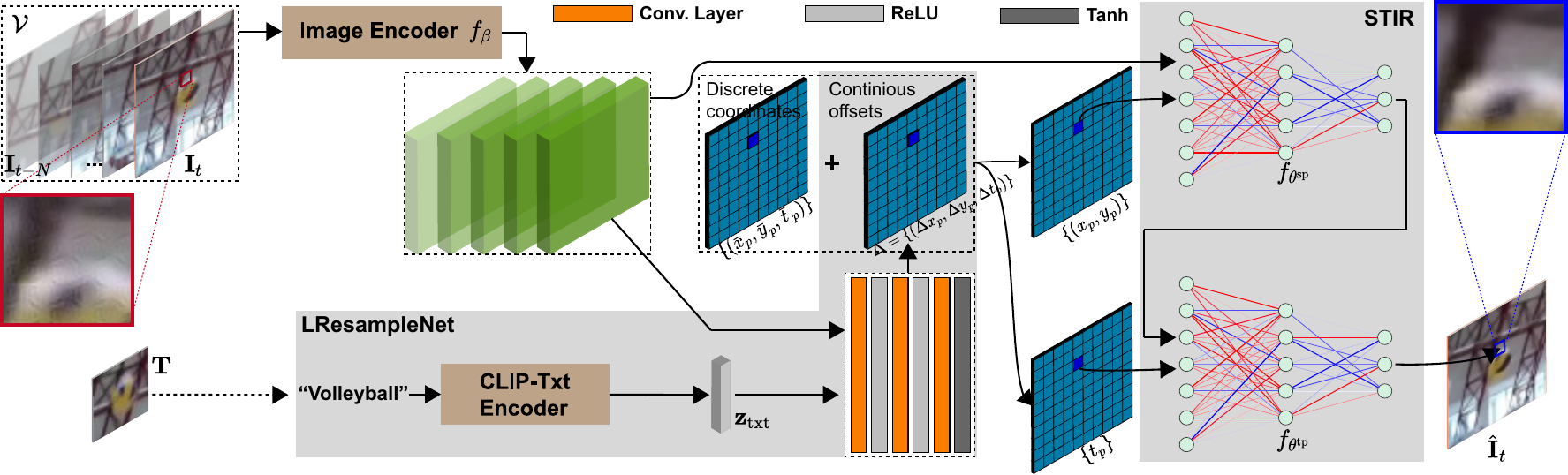}
\vspace{-10pt}
   \caption{Pipeline of proposed language-driven resamplable continuous representation (LRR) that contains two key parts, \ie, spatial-temporal implicit representation (STIR) and language-driven ResampleNet (LResampleNet). STIR takes continuous spatial and temporal coordinates as inputs (See point center at the blue rectangle) and estimates the corresponding color value.}
\label{fig:pipeline}
\vspace{-15pt}
\end{figure}

To address the first challenge, we propose to build a spatial-temporal implicit representation (See STIR in \Secref{subsec:ST_implicit}) that enables the reconstruction of any pixels at continuous spatial and temporal coordinates, which can remove the adversarial perturbations effectively and achieve appearance consistency to the clean counterpart \citep{chen2021learning,ho2022disco}. 
Regarding the second challenge, we propose a language-driven resample network (\ie, LResampleNet in \Secref{subsec:LResampleNet}) based on the built spatial-temporal implicit representation, which is able to generate a new frame by feeding resampled continuous coordinates to the STIR under the guidance of the text from the object template. 
Such a module makes the resampled frame have the same semantic text as the object template, naturally leading to semantic consistency. We display the whole pipeline in \Figref{fig:pipeline}.
%

%

\subsection{Spatial-Temporal Implicit Representation (STIR)}
\label{subsec:ST_implicit}

Given an image sequence $\mathcal{V}=\{\mathbf{I}_{\tau}\in\mathds{R}^{H\times W}\}_{\tau=t-N}^t$ containing the $t$th frame and its historical $N$ neighboring frames, we aim to construct an implicit representation for the sequence, \ie, $\hat{\mathcal{V}}$, which maps the spatial and temporal coordinates of a pixel (\ie, $\mathbf{p}=(x_p,y_p,\tau_p)$) in the continuous domain to the corresponding RGB value (\ie, $\hat{\mathcal{V}}(\mathbf{p})$). 
To this end, we propose to extend the recent local implicit image representation \citep{chen2021learning, ho2022disco} to the spatial-temporal domain.
In a straightforward way, we can formulate the task as
\begin{align}\label{eq:stir}
    \hat{\mathcal{V}}(\mathbf{p}) = \sum_{\mathbf{q}\in\mathcal{N}_\mathbf{p}} \omega_\mathbf{q} f_\theta(\mathbf{z}_\mathbf{q}, \text{dist}(\mathbf{p},\mathbf{q})),
\end{align}
where $\mathbf{p} = (x_p,y_p,\tau_p)$ is the coordinates of a pixel in the continuous spatial and temporal domain, that is, $x_p\in [0,H-1]$ and $y_p\in [0,W-1]$ can be non-integer and determines the spatial position of the pixel while $\tau_p\in [t-N,t]$ can be non-integer and decide its temporal location. 
The set $\mathcal{N}_\mathbf{p}$ contains neighboring pixels of the pixel $\mathbf{p}$ in $\mathcal{V}$. 
The vector $\mathbf{z}_\mathbf{q}$ denotes the feature of the pixel $\mathbf{q}$, and the function $\text{dist}(\mathbf{p},\mathbf{q})$ measures the spatial distance between the two pixels (\eg, Euclidean distance). 
The function $f_\theta$ is parameterized as an MLP. 
Intuitively, the function $f_\theta(\mathbf{z}_\mathbf{q}, \text{dist}(\mathbf{p},\mathbf{q}))$ is to map the feature of neighboring pixel $\mathbf{q}$ to the color of $\mathbf{p}$ based on their spatial distance. All generated color values are weightedly aggregated and the weight $\omega_\mathbf{q}$ is determined by the volume ratio of the cube formed by $\mathbf{p}$ and $\mathbf{q}$ to the total neighboring volume.

The complexity of the above formulation (\eg, \Eqref{eq:stir}) is directly related to the size of the neighboring set. For example, we consider $K\times K$ spatial neighboring pixels across the $N$ neighboring temporal frames. Then, the complexity of a pixel's reconstruction is $\mathcal{O}(NK^2)$. To alleviate the computing costs, we propose to decompose the reconstruction along the spatial and temporal domains and reformulate the \Eqref{eq:stir}. Specifically, we first build a spatial implicit representation that estimates the color values of a spatial location across all neighboring frames; that is, we have
\begin{align}\label{eq:stir_sp}
    \hat{\mathcal{V}}(\mathbf{p}_{(t-N:t)}) = \sum_{(x_q,y_q)\in\mathcal{N}_{(x_p,y_p)}} \omega_{(x_q,y_q)}^\text{sp}f_{\theta^\text{sp}}(\mathbf{z}_{\mathbf{q}_{(t-N:t)}}, \text{dist}((x_p,y_p),(x_q,y_q))),
\end{align}
where $\mathbf{p}_{(t-N:t)} = [(x_p,y_p,t-N),\ldots,(x_p,y_p,t)]$ and $\hat{\mathcal{V}}(\mathbf{p}_{(t-N:t)})$ preserves the $N$ color values of pixels at poistion $(x_p,y_p)$ across all temporal frames. 
The term $\mathbf{z}_{\mathbf{q}_{(t-N:t)}}$ concatenates the features of all pixels at location $(x_p,y_p)$ across all temporal frames. 
The function $f_{\theta^\text{sp}}$ is an MLP with the parameter $\theta^\text{sp}$, and the weight $\omega_{(x_q,y_q)}^\text{sp}$ is determined by the area ratio of the rectangle formed by $(x_p,y_p)$ and $(x_q,y_q)$ to the total neighboring areas as done in \citep{chen2021learning}.
After getting $\hat{\mathcal{V}}(\mathbf{p}_{(t-N:t)})$, we further build a temporal implicit representation that can estimate the color value of the pixel $\mathbf{p}=(x_p,y_p,\tau_p)$, that is, we have
\begin{align}\label{eq:stir_tp}
    \hat{\mathcal{V}}(\mathbf{p}) = \sum_{\tau_q\in [t-N,t]} \omega_{\tau_q}^\text{tp}f_{\theta^\text{tp}}(\hat{\mathcal{V}}(\mathbf{p}{(t-N:t)})[\tau_q], \text{dist}(\tau_p,\tau_q)),
\end{align}
where $\mathcal{V}(\mathbf{p}{(t-N:t)})[\tau_q]$ 
is the $\tau_q$th element in $\mathcal{V}(\mathbf{p}{(t-N:t)})$, and $f_{\theta^\text{tp}}(\cdot)$ is also an MLP to map the predicted $\mathcal{V}(\mathbf{p}{(t-N:t)})[\tau_q]$ to color value of the pixel $\mathbf{p}$. 
Compared with \Eqref{eq:stir}, the complexity of \Eqref{eq:stir_sp} and \Eqref{eq:stir_tp} is reduced to $\mathcal{O}(K^2+N)$.

We can simplify the developed STIR (\ie, \Eqref{eq:stir_sp} and \Eqref{eq:stir_tp}) as
\begin{align}\label{eq:stir_sim}
    \hat{\mathcal{V}}(\mathbf{p}) = \text{STIR}(\mathbf{p},\mathcal{V}|f_{\beta}, f_{\theta^\text{sp}}, f_{\theta^\text{tp}})
\end{align}
where $f_{\beta}$ is an encoder network to extract pixels' features (\ie, $z_q$ in \Eqref{eq:stir}). Once we train the parameters of $f_{\beta}$, $f_{\theta^\text{sp}}$, $f_{\theta^\text{tp}}$, we can generalize STIR to build implicit representations for arbitrary image sequences.

\subsection{Language-Driven ResampleNet (LResampleNet)}
\label{subsec:LResampleNet}

With the STIR, we can resample the $t$th frame by 
\begin{align}\label{eq:lresample}
    \hat{\mathbf{I}}_t(\bar{\mathbf{p}}) = \text{STIR}(\mathbf{p},\mathcal{V}|f_{\beta}, f_{\theta^\text{sp}}, f_{\theta^\text{tp}}), \text{with}~\mathbf{p} = \bar{\mathbf{p}} + \Delta \mathbf{p}
\end{align}
where $\bar{\mathbf{p}}=(\bar{x}_p, \bar{y}_p,t)$ is the discrete coordinate of the $t$th frame, that is,  $\bar{x}_p,$ and $\bar{y}_p$ are integers sampled from $[0,H-1]$ and $[0,W-1]$, respectively. 
Note that, we fix the temporal coordinate as $t$ since we handle the $t$th frame.
$\Delta \mathbf{p}$ are continuous offsets to generate continuous coordinates (\ie, $\mathbf{p}$) based on the integer coordinates $\bar{\mathbf{p}}$.
Hence, if we iterate through all discrete coordinates within the frame $\mathbf{I}_t$, we can reconstruct the $t$th frame and get $\hat{\mathbf{I}}_t$.
The key problem is how to predict the offset $\Delta \mathbf{p}$.
We propose to use the language extracted from the template $\mathbf{T}$ and the pixel's feature to guide the resampling, that is, to generate the offset for each pixel in $\mathbf{I}_t$.

Specifically, we initiate the process by employing CLIP \citep{radford2021learning}'s image encoder to extract the template (\ie, $\mathbf{T}$)'s embedding. Subsequently, given a set of texts encompassing potential categories of the object, we compare the template's embedding with the embeddings of all the texts. Following this, we select the text embedding that exhibits the highest similarity with the template's embedding as the $\mathbf{z}_\text{txt}$. Note that the text set can be updated based on different application scenarios, and alternative vision-language models or image caption methods can also be employed to achieve the same objective.
After that, we design a convolutional neural network denoted as \textit{ language-driven resampleNet (LResampleNet)} that takes the template's text embedding and pixel's feature embedding as inputs and predicts the offset; that is, we have
\begin{align} \label{eq:lresamplenet}
    \Delta = \text{LResampleNet}( \mathbf{Z}, \mathbf{z}_\text{txt})
\end{align}
where $\mathbf{Z}\in\mathds{R}^{H\times W\times C}$ contains the $C$-channel features of $HW$ pixels in $\mathbf{I}_t$ and is extracted via the encoder network $f_\beta(\cdot)$, and 
$\mathbf{z}_\text{txt}\in\mathds{R}^{1\times M}$ is the text embedding of the object template.  
In practice, we concatenate each pixel's feature with the text embedding and feed them to the \text{LResampleNet}. 
The output $\Delta \in \mathds{R}^{H\times W\times 3}$ contains the offsets of all pixels.

\subsection{Implementation Details}
\label{subsec:method-implementation}

\textbf{Architectures.} We set the $f_{\theta^\text{sp}}$ and $f_{\theta^\text{tp}}$ are five-layer MLPs with a ReLU activation layer and the hidden dimensions are 256. We use the network of \citep{lim2017enhanced} without the upsampling modules as the encoder for extracting pixel features (\ie, $f_{\beta}$), which can generate a feature with the same size as the input image.
\newline
\textbf{Loss function.} Given an attacked image sequence $\mathcal{V}=\{\mathbf{I}_\tau\}_{\tau=t-N}^t$ and the object template $\mathbf{T}$, we obtain the reconstructed $t$th frame $\hat{\mathbf{I}}_t$. When we have the clean version of $\hat{\mathbf{I}}_t$ (\ie, $\mathbf{I}_t^*$), we follow existing works and only use the $L_1$ loss function to train the STIR and LResampleNet. Intuitively, following the objectives in \Secref{subsec:method-overview}, we can add a consistency loss for the features of $\hat{\mathbf{I}}_t$ and $\mathbf{T}$ but we do not see clear benefits. 
\newline
\textbf{Training datasets.} We employ three widely-used datasets, \ie, ImageNet-DET \citep{russakovsky2015imagenet}, ImageNet-VID, and YouTube-BoundingBoxes \citep{real2017youtube} to train the STIR. 
Specifically, given a randomly sampled video, we randomly select five continuous frames in the video to form an image sequence and crop the object template $\mathcal{T}$ from another randomly chosen frame. 
Then, we add adversarial perturbations to the image sequence and regard the perturbed sequence as the $\mathcal{V}$ in \Eqref{eq:stir_sim}.
Here, we apply the FGSM attack on a pre-trained SiamRPN++ with ResNet50 tracker to produce adversarial perturbations.
After that, we have a pair of $\mathcal{V}$ and $\mathcal{T}$ as the training sample. 
We have sampled around 490,000 pairs for training STIR and LResampleNet, and 20,000 pairs as the validation set.
We train the STIR and LResampleNet independently since they have different functionalities, and joint training could hardly get good results for both modules. 
Besides, ImageNet-DET is an image dataset and we perform random translations on its images to get an image sequence to enlarge the training datasets.
\newline
\textbf{Other details.} We train and perform our method on a server with an NVIDIA RTX A6000 GPU and an Intel Core i9-10980XE 3.0GHz CPU using Pytorch \citep{paszke2019pytorch}. 
In alignment with the tracker's design, we have configured the reconstruction range to be the search region rather than the entire image, resulting in a significant reduction in time costs.
\newline
\textbf{LRR for adversarial tracking defense.}
LRR has a high generalization. After training LRR, we can use it to defend against diverse attacks for different trackers on any tracking dataset.
Specifically, given an incoming frame, we can employ the \Eqref{eq:lresample} and \Eqref{eq:lresamplenet} to reconstruct it and feed it to subsequent trackers to estimate the object's location and size.

\section{Experimental Results}

We conduct a series of experiments to evaluate LRR's defensive efficacy under various previously discussed settings, reporting the average results from three independent trials.
\newline
{\bf Testing datasets.}
For evaluate the effectiveness of adversarial defense approach, we utilized three widely used tracking datasets: OTB100 \citep{wu_object_2015_otb100}, VOT2019 \citep{vot2019}, and UAV123 \citep{mueller_benchmark_2016_uav123}.
VOT2019 and OTB100 are popular tracking datasets that consist of 60 and 100 videos, respectively. 
UAV123 dataset focuses on object tracking in videos captured by uncrewed aerial vehicle cameras, containing 123 videos. 
\newline
{\bf Trackers and attacks.}
Given the variance in adversarial attacks on VOT tasks across both algorithms and implementations, it is crucial to employ representative trackers to facilitate a comprehensive and impartial assessment of adversarial attack resilience. 
This approach also serves to demonstrate the general efficacy of our proposed defense mechanism. 
To this end, we select trackers from the SiamRPN++ family: SiamRPN++ with ResNet50 and SiamRPN++ with MobileNetV2, and identify four challenging attackers, the IoU Attack \citep{jia2021iou}, SPARK \citep{guo2020spark}, CSA \citep{yan2020cooling}, and RTAA \citep{jia2020robust}, which are known to deliver robust performance against SiamRPN++ trackers. We detail the implementations of these attacks in \ref{app:trackingattacks}.
\newline
{\bf Defence baselines.}
To assess the effectiveness of our proposed method comprehensively, we compare it against adversarial fine-tuning techniques and SOTA adversarial defense approach. 
Adversarial fine-tuning, as outlined by \citep{goodfellow2014explaining}, is a strategy that trains a model with both clean and adversarial examples, thereby enhancing the model's resilience against attacks. 
For the adversarial fine-tuning baseline, we employ FGSM \citep{goodfellow2014explaining}, PGD \citep{madry2017towards}, and CSA \citep{yan2020cooling} to generate adversarial examples and augment the training data, thereby enabling the model to fortify its defenses against adversarial attacks. 
Both PGD and FGSM add minimal calculated perturbation to the input image based on the gradient of the tracker model's loss concerning the input, while CSA uses its perturbation generator to inject adversarial examples, progressively reducing the confidence of the tracker's backbone. 
For the adversarial defense method, we adapt the SOTA method, DISCO \citep{ho2022disco}, for tracking tasks, using it to predict each pixel's RGB value through local implicit functions, thus defending against attacks.
We incorporate DISCO as a frame processor into our adversarial tracking attacks defense task.

\subsection{Comparison Results}

\begin{table}
\centering
\setlength{\extrarowheight}{0pt}
\addtolength{\extrarowheight}{\aboverulesep}
\addtolength{\extrarowheight}{\belowrulesep}
\setlength{\aboverulesep}{0pt}
\setlength{\belowrulesep}{0pt}
\caption{Comparing LRR with baselines on OTB100, VOT2019, and UAV123 under Four Attacks.}
\label{tab:overall_results}
\resizebox{0.9\linewidth}{!}{%
\begin{tabular}{c|c|ccccc|ccccc|ccccc} 
\toprule
\multirow{2}{*}{SiamRPN++} & \multirow{2}{*}{Defenses} & \multicolumn{5}{c|}{OTB100 Prec. (\%)} & \multicolumn{5}{c|}{VOT2019 EAO} & \multicolumn{5}{c}{UAV123 Prec. (\%)} \\
 &  & Cln. & RTAA & IoU & CSA & SPARK & Cln. & RTAA & IoU & CSA & SPARK & Cln. & RTAA & IoU & CSA & SPARK \\ 
\hline
\multicolumn{1}{l|}{} & \multicolumn{1}{l|}{} & \multicolumn{1}{l}{} & \multicolumn{1}{l}{} & \multicolumn{1}{l}{} & \multicolumn{1}{l}{} & \multicolumn{1}{l|}{} & \multicolumn{1}{l}{} & \multicolumn{1}{l}{} & \multicolumn{1}{l}{} & \multicolumn{1}{l}{} & \multicolumn{1}{l|}{} & \multicolumn{1}{l}{} & \multicolumn{1}{l}{} & \multicolumn{1}{l}{} & \multicolumn{1}{l}{} & \multicolumn{1}{l}{} \\
\multirow{6}{*}{Res50} & wo.Def & \textbf{91.4} & 32.7 & 75.9 & 47.2 & 69.8 & \textbf{0.277} & 0.080 & 0.153 & 0.089 & 0.079 & \textbf{79.5} & 41.2 & 70.5 & 46.5 & 40.8 \\
 & AT$_\text{FGSM}$ & 85.1 & 53.5 & 77.6 & 60.7 & 69.6 & 0.214 & 0.100 & 0.176 & 0.109 & 0.078 & 77.4 & 48.9 & 72.5 & 44.9 & 41.3 \\
 & AT$_\text{PGD}$ & 81.8 & 50.2 & 76.8 & 62.3 & 61.0 & 0.218 & 0.090 & 0.171 & 0.125 & 0.057 & 79.4 & 55.6 & 74.0 & 68.8 & 37.0 \\
 & AT$_\text{CSA}$ & 84.3 & 52.2 & 77.2 & 80.9 & 65.1 & 0.251 & 0.090 & 0.164 & 0.152 & 0.072 & 76.8 & 43.2 & 70.6 & 74.2 & 34.5 \\
 & DISCO & 86.0 & 86.3 & 78.6 & 83.6 & 85.7 & 0.249 & 0.244 & 0.190 & 0.204 & 0.248 & 79.1 & 76.8 & 75.9 & 77.7 & 76.0 \\
 & {\cellcolor[rgb]{0.878,0.882,1}}LRR & {\cellcolor[rgb]{0.878,0.882,1}}87.8 & {\cellcolor[rgb]{0.878,0.882,1}}\textbf{86.9} & {\cellcolor[rgb]{0.878,0.882,1}}\textbf{85.3} & {\cellcolor[rgb]{0.878,0.882,1}}\textbf{89.4} & {\cellcolor[rgb]{0.878,0.882,1}}\textbf{89.3} & {\cellcolor[rgb]{0.878,0.882,1}}0.262 & {\cellcolor[rgb]{0.878,0.882,1}}\textbf{0.255} & {\cellcolor[rgb]{0.878,0.882,1}}\textbf{0.217} & {\cellcolor[rgb]{0.878,0.882,1}}\textbf{0.237} & {\cellcolor[rgb]{0.878,0.882,1}}\textbf{0.269} & {\cellcolor[rgb]{0.878,0.882,1}}79.3 & {\cellcolor[rgb]{0.878,0.882,1}}\textbf{77.7} & {\cellcolor[rgb]{0.878,0.882,1}}\textbf{78.6} & {\cellcolor[rgb]{0.878,0.882,1}}\textbf{81.8} & {\cellcolor[rgb]{0.878,0.882,1}}\textbf{79.3} \\ 
\hline
\multirow{6}{*}{MobileV2} & wo.Def & 85.5 & 25.6 & 67.8 & 40.9 & 32.2 & \textbf{0.267} & 0.062 & 0.125 & 0.083 & 0.037 & \textbf{80.3} & 39.3 & 66.2 & 42.2 & 22.5 \\
 & AT$_\text{FGSM}$ & 79.4 & 35.6 & 72.6 & 63.1 & 30.1 & 0.213 & 0.070 & 0.144 & 0.121 & 0.041 & 78.4 & 39.3 & 67.1 & 64.1 & 21.6 \\
 & AT$_\text{PGD}$ & 77.7 & 34.9 & 72.0 & 73.0 & 23.9 & 0.202 & 0.075 & 0.122 & 0.130 & 0.035 & 77.9 & 36.7 & 67.7 & 71.3 & 18.7 \\
 & AT$_\text{CSA}$ & 79.4 & 34.2 & 60.2 & 76.6 & 35.7 & 0.263 & 0.078 & 0.097 & 0.137 & 0.037 & 74.8 & 43.4 & 56.4 & 70.9 & 14.8 \\
 & DISCO & 82.9 & 78.7 & 75.0 & 79.9 & 80.1 & 0.175 & 0.161 & 0.132 & 0.166 & 0.208 & 74.6 & 76.2 & 73.3 & 72.9 & 75.3 \\
 & {\cellcolor[rgb]{0.878,0.882,1}}LRR & {\cellcolor[rgb]{0.878,0.882,1}}\textbf{85.6} & {\cellcolor[rgb]{0.878,0.882,1}}\textbf{83.2} & {\cellcolor[rgb]{0.878,0.882,1}}\textbf{82.1} & {\cellcolor[rgb]{0.878,0.882,1}}\textbf{83.9} & {\cellcolor[rgb]{0.878,0.882,1}}\textbf{85.4} & {\cellcolor[rgb]{0.878,0.882,1}}0.240 & {\cellcolor[rgb]{0.878,0.882,1}}\textbf{0.205} & {\cellcolor[rgb]{0.878,0.882,1}}\textbf{0.166} & {\cellcolor[rgb]{0.878,0.882,1}}\textbf{0.223} & {\cellcolor[rgb]{0.878,0.882,1}}\textbf{0.239} & {\cellcolor[rgb]{0.878,0.882,1}}79.1 & {\cellcolor[rgb]{0.878,0.882,1}}\textbf{78.7} & {\cellcolor[rgb]{0.878,0.882,1}}\textbf{75.9} & {\cellcolor[rgb]{0.878,0.882,1}}\textbf{76.2} & {\cellcolor[rgb]{0.878,0.882,1}}\textbf{79.0} \\
\bottomrule
\end{tabular}
}
\vspace{-20pt}
\end{table}

LRR achieves SOTA performance over the baselines, as detailed in \tableref{tab:overall_results}, which analyzes adversarial defense under four attacks across three datasets and two SiamRPN++ family trackers. 
The LRR setup follows the approach in \Secref{subsec:method-implementation}. The table illustrates that SiamRPN++ trackers can be compromised, impacting precision on OTB100 and UAV123 and Expected Average Overlap (EAO) on VOT2019. 
FGSM and PGD, as adversarial fine-tuning approaches, provide minimal defense, decreasing performance even on non-attacked inputs. 
While CSA fine-tuning improves defense against its generator’s examples, it underperforms under other attacks. 
Overall, the adversarial fine-tuning baselines present a marginally successful defense against IoU and CSA but are ineffective against RTAA and SPARK.
Meanwhile, DISCO displays robust defense against all attack types but is outperformed by LRR due to its inability to leverage information between frames.
To validate the effectiveness further, we compare the visualizations of DISCO and LRR at both the image level and the response map level in the supplementary material \ref{app:extra-visualizations}. The results demonstrate that LRR can achieve higher consistency at the semantic and image quality levels than DISCO.

\subsection{Ablation Study and Discussion}
\label{subsec:ablation-and-discussion}

\begin{figure}[t]
\centering
\includegraphics[width=0.9\linewidth]{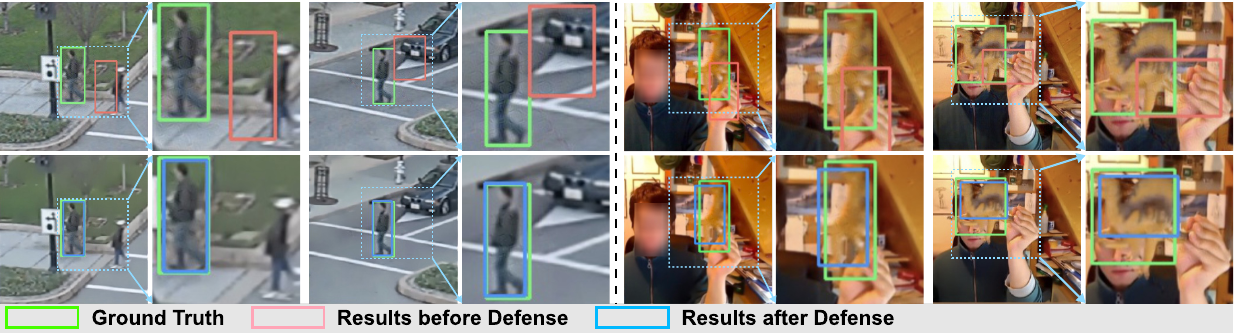}
\vspace{-10pt}
   \caption{Visualization comparison before \& after LRR defense for SiamRPN++ under CSA attack.}
\label{fig:vis}
\vspace{-10pt}
\end{figure}

%
\begin{table}\centering
\caption{Comparison between STIR alone and LResampleNet on VOT2019, OTB100, and UAV123.}
\label{tab:resample}
\resizebox{0.9\linewidth}{!}{%
\begin{tabular}{c|c|ccc|ccc|ccc} 
\toprule
\multirow{2}{*}{SiamRPN++} & \multirow{2}{*}{Attacks} & \multicolumn{3}{c|}{OTB100 Prec.~(\%)} & \multicolumn{3}{c|}{VOT2019 EAO} & \multicolumn{3}{c}{UAV123 Prec.~(\%)} \\
 &  & Org. & wo.LResample & LResample~ & Org. & wo.LResample~~ & LResample & Org. & wo.LResample~ & LResample~ \\ 
\hline
\multicolumn{1}{l|}{} & \multicolumn{1}{l|}{} & \multicolumn{1}{l}{} & \multicolumn{1}{l}{} & \multicolumn{1}{l|}{} & \multicolumn{1}{l}{} & \multicolumn{1}{l}{} & \multicolumn{1}{l|}{} & \multicolumn{1}{l}{} & \multicolumn{1}{l}{} & \multicolumn{1}{l}{} \\
\multirow{5}{*}{Res50} & wo.Atk & 91.4 & \textbf{88.1} & 87.8 & 0.277 & \textbf{0.268} & 0.262 & 79.5 & \textbf{79.5} & 79.3 \\
 & RTAA & 32.7 & 85.9 & \textbf{86.9} & 0.080 & 0.247 & \textbf{0.255} & 41.2 & 77.0 & \textbf{77.7} \\
 & IoUAttack & 75.9 & 84.0 & \textbf{85.3} & 0.153 & 0.213 & \textbf{0.217} & 70.5 & 78.2 & \textbf{78.6} \\
 & CSA & 47.2 & 85.9 & \textbf{89.4} & 0.089 & 0.219 & \textbf{0.237} & 46.5 & 79.7 & \textbf{81.8} \\
 & SPARK & 69.8 & 87.7 & \textbf{89.3} & 0.079 & 0.256 & \textbf{0.269} & 40.8 & 77.9 & \textbf{79.3} \\ 
\hline
\multirow{5}{*}{MobileV2} & wo.Atk & 85.5 & 85.5 & \textbf{85.6} & 0.267 & 0.238 & \textbf{0.240} & 80.3 & 78.4 & \textbf{79.1} \\
 & RTAA & 25.6 & 80.7 & \textbf{83.2} & 0.062 & 0.204 & \textbf{0.205} & 39.3 & 77.7 & \textbf{78.7} \\
 & IoUAttack & 67.8 & 79.0 & \textbf{82.1} & 0.125 & 0.163 & \textbf{0.166} & 66.2 & 74.9 & \textbf{75.9} \\
 & CSA & 40.9 & 79.8 & \textbf{83.9} & 0.083 & 0.216 & \textbf{0.223} & 42.2 & 75.7 & \textbf{76.2} \\
 & SPARK & 32.2 & 83.7 & \textbf{85.4} & 0.037 & 0.225 & \textbf{0.239} & 22.5 & 78.2 & \textbf{79.0} \\
\bottomrule
\end{tabular}
\vspace{-15pt}
}
\end{table}

\begin{table}
\centering
\vspace{-20pt}
\caption{Comparison of ResampleNet with \& without language guidance on three datasets.}
\label{tab:lang_guidance}
\resizebox{0.8\linewidth}{!}{%
\begin{tabular}{c|c|ccc|ccc|ccc} 
\toprule
\multirow{2}{*}{SiamRPN++} & \multirow{2}{*}{Attacks} & \multicolumn{3}{c|}{OTB100 Prec.~(\%)} & \multicolumn{3}{c|}{VOT2019 EAO} & \multicolumn{3}{c}{UAV123 Prec.~(\%)} \\
 &  & Org. & wo.Lang. & Lang.~ & Org. & wo.Lang.~ & Lang. & Org. & wo.Lang.~ & Lang.~ \\ 
\hline
\multicolumn{1}{l|}{} & \multicolumn{1}{l|}{} & \multicolumn{1}{l}{} & \multicolumn{1}{l}{} & \multicolumn{1}{l|}{} & \multicolumn{1}{l}{} & \multicolumn{1}{l}{} & \multicolumn{1}{l|}{} & \multicolumn{1}{l}{} & \multicolumn{1}{l}{} & \multicolumn{1}{l}{} \\
\multirow{5}{*}{Res50} & wo.Atk & 91.4 & \textbf{88.1} & 87.8 & 0.277 & \textbf{0.270} & 0.262 & 79.5 & \textbf{79.6} & 79.3 \\
 & RTAA & 32.7 & 86.1 & \textbf{86.9} & 0.080 & 0.250 & \textbf{0.255} & 41.2 & 77.4 & \textbf{77.7} \\
 & IoUAttack & 75.9 & 84.4 & \textbf{85.3} & 0.153 & \textbf{0.219} & 0.217 & 70.5 & 78.4 & \textbf{78.6} \\
 & CSA & 47.2 & 86.0 & \textbf{89.4} & 0.089 & 0.222 & \textbf{0.237} & 46.5 & 79.8 & \textbf{81.8} \\
 & SPARK & 69.8 & 87.8 & \textbf{89.3} & 0.079 & 0.254 & \textbf{0.269} & 40.8 & 78.1 & \textbf{79.3} \\ 
\hline
\multirow{5}{*}{MobileV2} & wo.Atk & 85.5 & \textbf{86.1} & 85.6 & 0.267 & 0.238 & \textbf{0.240} & 80.3 & \textbf{79.6} & 79.1 \\
 & RTAA & 25.6 & 81.2 & \textbf{83.2} & 0.062 & 0.201 & \textbf{0.205} & 39.3 & 78.0 & \textbf{78.7} \\
 & IoUAttack & 67.8 & 79.0 & \textbf{82.1} & 0.125 & 0.151 & \textbf{0.166} & 66.2 & 75.3 & \textbf{75.9} \\
 & CSA & 40.9 & 81.1 & \textbf{83.9} & 0.083 & 0.219 & \textbf{0.223} & 42.2 & 76.0 & \textbf{76.2} \\
 & SPARK & 32.2 & 84.5 & \textbf{85.4} & 0.037 & 0.228 & \textbf{0.239} & 22.5 & 78.4 & \textbf{79.0} \\
\bottomrule
\end{tabular}
}
\vspace{-5pt}
\end{table}

In this section, we explore the effectiveness of the components in our LRR (Language-Driven Resampling Network), specifically discussing the individual contributions of the resampling network, language-driven approach, and spatial-temporal information toward the defense mechanism.
\newline
{\bf Overall results.}
LRR has demonstrated robust defense against adversarial attacks. 
Employing the VOT2019’s Expected Average Overlap (EAO) metric, a composite measure of Accuracy and Robustness, it is evident from \tableref{tab:overall_results} that our defenses significantly enhanced EAO. 
Following the implementation of our defense, the average EAO value under attack increased to 89\% and 81\% for the SiamRPN++ with ResNet50 and MobileNetV2 trackers, respectively. 
Additionally, using precision as a metric for the OTB100 and UAV123 datasets, our defense approach has shown a boost in precision to 90\% across all attackers and trackers, highlighting its effectiveness.
Furthermore, we extend our evaluation to four additional widely used datasets, including one large-scale dataset, as detailed in \ref{app:extra-datasets}.
This extended evaluation demonstrates the effective transferability of our method across diverse datasets.
In \ref{app:baseline_resize_compress}, we also compare with image resizing and compression-based defense methods, which further demonstrates the advantages of our method.
\newline
{\bf Illustrative Overview of Defense Results.}
\Figref{fig:vis} qualitatively demonstrates the defense results achieved by LRR. Our method removes the adversarial textures effectively and makes the tracker localize the object of interest accurately.
In \ref{app:extra-visualizations}, we delve further into visualizations on correlation maps and discuss in greater depth the impact of our method on adversarial defense.
\newline
{\bf Effectiveness of resampling.}
To validate the effectiveness of our primary contribution, we conducted experiments to demonstrate the influence of the LResampleNet in LRR.
Given STIR's independent training with LResampleNet, it should estimate perturbations utilizing spatial-temporal information.
We evaluated STIR without resampling, following the experimental settings of previous evaluations. 
In \tableref{tab:resample}, we present the increase in precision for the OTB100 and UAV123 datasets and the rise in EAO value for VOT2019. 
In \tableref{tab:resample}, results indicate that tracking outcomes without the LResampleNet are less effective than LRR in defending against adversarial tracking attacks. A more detailed discussion on this is articulated in \ref{app:resampling-discussion}.
\newline
{\bf Effectiveness of language guidance.}
When introducing the resampling mechanism into our pipeline in \Secref{subsec:method-implementation}, we used language to establish a connection between incoming frames and the tracking template, constituting a major contribution to our work. 
Since we feed both pixel embedding and text embedding to the resampling network, we aim to validate the effectiveness of our language-driven approach. 
We designed a resampling network without text embedding (ResampleNet), allowing pixel embedding to serve as the sole input, replacing the LResampleNet in our existing pipeline.
As shown in \tableref{tab:lang_guidance}, the use of ResampleNet guidance appears to be less effective when compared to our LRR pipeline. 
However, compared to the pipeline that uses STIR alone, ResampleNet demonstrates an enhanced defense against adversarial tracking attacks. 
The primary reason for this is ResampleNet's ability to estimate adversarial perturbations by leveraging the implicit continuous representation from the input pixel embedding.
\begin{table}
\centering
\caption{Comparison of STIR with different settings of $N$, evaluated on OTB100 and VOT2019.}
\label{tab:spatial_info}
\resizebox{0.9\linewidth}{!}{%
\begin{tabular}{c|c|cccccc|cccccc} 
\toprule
\multirow{2}{*}{SiamRPN++} & \multirow{2}{*}{Attacks} & \multicolumn{6}{c|}{OTB100 Prec.~(\%)} & \multicolumn{6}{c}{VOT2019 EAO} \\
 &  & Org. & $N=1$ & $N=2$ & \multicolumn{1}{l}{$N=3$} & \multicolumn{1}{l}{$N=4$} & \multicolumn{1}{l|}{$N=5$} & Org. & $N=1$ & $N=2$ & \multicolumn{1}{l}{$N=3$} & \multicolumn{1}{l}{$N=4$} & \multicolumn{1}{l}{$N=5$} \\ 
\hline
\multicolumn{1}{l|}{} & \multicolumn{1}{l|}{} & \multicolumn{1}{l}{} & \multicolumn{1}{l}{} & \multicolumn{1}{l}{} & \multicolumn{1}{l}{} & \multicolumn{1}{l}{} & \multicolumn{1}{l|}{} & \multicolumn{1}{l}{} & \multicolumn{1}{l}{} & \multicolumn{1}{l}{} & \multicolumn{1}{l}{} & \multicolumn{1}{l}{} & \multicolumn{1}{l}{} \\
\multirow{5}{*}{Res50} & wo.Atk & 91.4 & 86.5 & 87.3 & 87.1 & 87.3 & 88.1 & 0.277 & 0.237 & 0.251 & 0.255 & 0.267 & 0.268 \\
 & RTAA & 32.7 & 84.9 & 85.0 & 85.6 & 85.6 & 85.9 & 0.080 & 0.240 & 0.241 & 0.241 & 0.245 & 0.247 \\
 & IoUAttack & 75.9 & 78.7 & 79.2 & 80.9 & 83.1 & 84.0 & 0.153 & 0.190 & 0.191 & 0.208 & 0.211 & 0.213 \\
 & CSA & 47.2 & 82.6 & 82.8 & 83.3 & 85.2 & 85.9 & 0.089 & 0.203 & 0.204 & 0.215 & 0.216 & 0.219 \\
 & SPARK & 69.8 & 85.5 & 85.8 & 86.3 & 87.0 & 87.7 & 0.079 & 0.245 & 0.249 & 0.251 & 0.252 & 0.256 \\
\bottomrule
\end{tabular}
}
\vspace{-15pt}
\end{table}

\newline
{\bf Effectiveness of spatial-temporal information.}
To validate STIR in learning spatial-temporal information, we trained it separately by altering the input frame length $N\in\{1,2,3,4,5\}$ from the training dataset described in \Secref{subsec:method-implementation}. 
To assess the influence of LResampleNet, we evaluated these STIR models independently without the integration of our LRR on OTB100 and VOT2019 datasets using SiamRPN++ with ResNet50 tracker.
The results presented in \tableref{tab:spatial_info}, reveal that as the number of frame inputs length ${N}$ increases, STIR demonstrates an enhanced capability to defend against adversarial tracking attacks.
This suggests that STIR is able to extract more hidden information from spatial-temporal information brought by input frames, thereby serving a better purpose in constructing video frame RGB values from it.
\newline
{\bf Transferability to transformer-based trackers.}
To clarify the transferability of our LRR approach, we adapted our method to the recently proposed transformer-based tracker model, ToMP-50 \citep{mayer2022transforming}.
\begin{wraptable}{h}{0.5\textwidth}
\centering
\vspace{-15pt}
\caption{Defense on ToMP-50 across 3 datasets}
\label{tab:appendix_tomp}
\resizebox{1.0\linewidth}{!}{%
\setlength\tabcolsep{2pt}
\begin{tabular}{c|c|cc|cc|cc} 
\toprule
\multirow{2}{*}{ToMP} & \multirow{2}{*}{Attacks} & \multicolumn{2}{c|}{OTB100 Prec.~(\%)} & \multicolumn{2}{c|}{VOT2019 EAO} & \multicolumn{2}{c}{UAV123 Prec.~(\%)} \\
 &  & Org. & LRR & Org. & LRR & Org. & LRR \\ 
\hline
\multicolumn{1}{l|}{} & \multicolumn{1}{l|}{} & \multicolumn{1}{l}{} & \multicolumn{1}{l|}{} & \multicolumn{1}{l}{} & \multicolumn{1}{l|}{} & \multicolumn{1}{l}{} & \multicolumn{1}{l}{} \\
\multirow{2}{*}{ToMP-50} & wo.Atk & 90.1 & \textbf{89.8} & 0.556 & \textbf{0.547} & 88.2 & \textbf{87.8} \\
 & RTAA & 61.8 & \textbf{90.0} & 0.337 & \textbf{0.552} & 58.5 & \textbf{88.0} \\
\bottomrule
\end{tabular}
}
\vspace{-12pt}
\end{wraptable}

Specifically, we employed RTAA to attack ToMP-50 and applied our LRR method for defense, evaluating the results across three different datasets. 
The results, delineated in \tableref{tab:spatial_info}, underscore the transferability of our proposed method, sustaining its efficacy even when incorporated with newly developed tracking models.
A detailed discussion can be found in \ref{app:transformer-based-tracker}.
\newline
{\bf Defense efficiency.}
LRR addresses attacks via the elimination of perturbations at testing time.
This strategy allows our method to be easily integrated into various existing tracking task pipelines, which also raises the concern of additional computational consumption.
\begin{wraptable}{h}{0.5\textwidth}
\centering
\vspace{-15pt}
\caption{Average time costs on OTB100.}
\label{tab:time_cost}
\resizebox{1.0\linewidth}{!}{%
\begin{tabular}{c|c|cccc|cc} 
\toprule
\multirow{3}{*}{SiamRPN++} & \multicolumn{7}{c}{Cost per frame (ms)} \\
\cline{2-8}
 & Track & \multicolumn{4}{c|}{Attack} & \multicolumn{2}{c}{Defense} \\
 & - & RTAA & IoUAttack & CSA & SPARK & STIR & LRR \\ 
\hline
\multicolumn{1}{l|}{} & \multicolumn{1}{l|}{} & \multicolumn{1}{l}{} & \multicolumn{1}{l}{} & \multicolumn{1}{l}{} & \multicolumn{1}{l|}{} & \multicolumn{1}{l}{} & \multicolumn{1}{l}{} \\
Res50 & 16 & 215 & 1184 & 4 & 76 & 34 & 39 \\
MobileV2 & 13 & 118 & 667 & 4 & 62 & 34 & 39 \\
\bottomrule
\end{tabular}
}
\vspace{-15pt}
\end{wraptable}

We report the time cost of our methods in \tableref{tab:time_cost}.
Using our proposed method as a standalone frame processor, our STIR defense can maintain processing at approximately 29 fps. 
In comparison, LRR operates at around 25 fps. 
This allows for the facilitation of online tracking adversarial defense capability.
For a more detailed discussion, please refer to \ref{app:defense-efficiency}.

\section{Conclusion}

In this work, we have developed a novel implicit representation, \ie, language-driven resamplable continuous representation (LRR), against state-of-the-art adversarial tracking attacks. We first built a spatial-temporal implicit representation (STIR) to utilize the spatial-temporal neighboring pixels for effective appearance reconstruction. Then, we designed the language-driven ResampleNet to encourage semantic consistency between the reconstructed frame and the object template.
After training on large-scale datasets, our method can be used to defend against different attacks for different trackers on different testing datasets.
Impressively, our method has successfully defended four state-of-the-art attacks and let the adversarial accuracy approach the clean accuracy while maintaining the high accuracy on the clean data. 
\newline
\textbf{Limitations.} As an extra module, the proposed method inevitably increases the computing and time costs. In the future, we can explore approaches to decrease costs. Besides, the generalization to non-noise-based attacks like motion blur \citep{guo2021learning} should be future studied. 
Furthermore, in recent years, researchers have increasingly directed their attention toward natural language-specified visual object tracking \citep{wang2021towards}, which offers greater flexibility in real-world scenarios. However, existing attack and defense methods predominantly focus on template-based trackers, overlooking this emerging trend. Future research endeavors should aim to bridge this gap.

\newpage
\section{Reproducibility Statement}

To facilitate the reproducibility of our approach, we have open-sourced our code and provided a benchmark that includes our method, which is accessible via \url{https://github.com/tsingqguo/robustOT}. 
This repository contains the essential evaluation code, along with comprehensive instructions to facilitate the deployment of the proposed methods and the establishment of the evaluation environment. 
The repository also includes a pre-trained model, allowing for direct replication of the demonstrated results.

All implementation details are meticulously described in \Secref{subsec:method-implementation}. 
The thorough documentation, along with the availability of the benchmark and pre-trained model, aims to assist in the validation and replication of the presented results.

\section*{Acknowledgment}

This research is supported by the National Research Foundation, Singapore, and DSO National Laboratories under the AI Singapore Programme (AISG Award No: AISG2-GC-2023-008), and Career Development Fund (CDF) of Agency for Science, Technology and Research (A*STAR) (No.: C233312028).
This work is supported in part by funding from the Canada First Research Excellence Fund as part of the University of Alberta’s Future Energy Systems research initiative, Canada CIFAR AI Chairs Program, the Natural Sciences and Engineering Research Council of Canada (NSERC No.RGPIN-2021-02549, No.RGPAS-2021-00034, No.DGECR-2021-00019); as well as JST-Mirai Program Grant No.JPMJMI20B8, JSPS KAKENHI Grant No.JP21H04877, No.JP23H03372, and the support from TIER IV, Inc. and Autoware Foundation.

\newpage

\bibliography{iclr2024_conference}
\bibliographystyle{iclr2024_conference}

\newpage
\appendix
\section{Appendix}

\subsection{Transferability across Datasets}
\label{app:extra-datasets}

In this section, we extend the evaluation of our defense approach, LRR, to additional datasets to investigate its transferability. 
Specifically, we test LRR's performance on the challenging LaSOT \citep{fan2019lasot}, NFS \citep{kiani2017need} and TrackingNet \citep{muller2018trackingnet} datasets.
The LaSOT is a large-scale dataset containing 280 videos. 
On the other hand, the NFS dataset consists of 100 videos captured using a high-speed camera and is divided into two variants: NFS30 and NFS240, which have frame rates of 30 fps and 240 fps, respectively.
Additionally, TrackingNet encompasses a diverse set of 511 video sequences, offering a broad range of real-world scenarios to rigorously evaluate tracking algorithms.

\begin{table}[h]
\centering
\caption{Results of LRR defense over four extended datasets.}
\label{tab:appendix_overall_results}
\resizebox{0.8\linewidth}{!}{%
\begin{tabular}{c|c|cc|cc|cc|cc}
\toprule
\multirow{2}{*}{SiamRPN++} & \multirow{2}{*}{Attacks} & \multicolumn{2}{c|}{LaSOT Prec.~(\%)} & \multicolumn{2}{c|}{NFS30 Prec.~(\%)} & \multicolumn{2}{c|}{NFS240 Prec.~(\%)} & \multicolumn{2}{c}{TrackingNet Prec.~(\%)} \\
 &  & Org. & LRR & Org. & LRR & Org. & LRR & Org. & LRR \\ 
\hline
\multicolumn{1}{l|}{} & \multicolumn{1}{l|}{} & \multicolumn{1}{l}{} & \multicolumn{1}{l|}{} & \multicolumn{1}{l}{} & \multicolumn{1}{l|}{} & \multicolumn{1}{l}{} & \multicolumn{1}{l|}{} &  &  \\
\multirow{5}{*}{Res50} & wo.Atk & \textbf{48.8} & 48.7 & \textbf{59.9} & 56.0 & \textbf{71.2} & 69.9 & \textbf{69.4} & 67.7 \\
 & RTAA & 20.5 & \textbf{46.3} & 22.4 & \textbf{56.5} & 37.4 & \textbf{69.6} & 13.9 & \textbf{65.7} \\
 & IoUAttack & 39.6 & \textbf{46.5} & 42.0 & \textbf{55.5} & 65.3 & \textbf{67.8} & 61.9 & \textbf{65.9} \\
 & CSA & 17.5 & \textbf{45.3} & 19.6 & \textbf{58.0} & 33.5 & \textbf{70.5} & 39.7 & \textbf{66.5} \\
 & SPARK & 19.6 & \textbf{48.5} & 40.5 & \textbf{59.3} & 16.2 & \textbf{70.6} & 29.6 & \textbf{55.1} \\ 
\hline
\multirow{5}{*}{MobileV2} & wo.Atk & \textbf{44.8} & 44.0 & \textbf{57.2} & 55.8 & \textbf{69.0} & 66.7 & 63.6 & \textbf{63.9} \\
 & RTAA & 12.5 & \textbf{40.7} & 16.8 & \textbf{55.0} & 25.3 & \textbf{68.1} & 5.2 & \textbf{62.5} \\
 & IoUAttack & 29.6 & \textbf{41.4} & 30.7 & \textbf{47.8} & 55.8 & \textbf{66.3} & 54.1 & \textbf{60.8} \\
 & CSA & 11.3 & \textbf{37.8} & 17.6 & \textbf{55.6} & 21.4 & \textbf{64.7} & 30.2 & \textbf{59.5} \\
 & SPARK & 10.2 & \textbf{43.9} & 19.5 & \textbf{56.8} & 7.0 & \textbf{66.5} & 17.1 & \textbf{54.1} \\
\bottomrule
\end{tabular}
}
\end{table}

From \tableref{tab:appendix_overall_results}, we can observe that our LRR exhibits excellent transferability over large-scale datasets. It successfully defends against adversarial tracking attacks across these challenging datasets.

\subsection{Detailed Analysis of LResampleNet's Impact}
\label{app:resampling-discussion}

To validate the effectiveness of our primary contribution, we conducted experiments to demonstrate the influence of the LResampleNet in LRR.
Given the independent training of STIR and LResampleNet, STIR should be capable of estimating the perturbation by using spatial-temporal information.
We evaluated STIR without resampling and assessed performance on clean data and four attackers across three datasets on two trackers from the SiamRPN++ family. 
In \tableref{tab:resample}, we present the increase in precision for the OTB100 and UAV123 datasets and the rise in EAO value for VOT2019. 
The results indicate that tracking outcomes without the LResampleNet are less effective than LRR in defending against adversarial tracking attacks.
However, it has been observed that using STIR alone causes less damage to the clean data when compared to the LRR approach. 
This suggests that LRR has the potential to damage clean data. 
Nevertheless, considering the overall results, an accuracy of less than $2\%$ for OTB100 and UAV123, or an EAO of $0.01$ for VOT2019 can be deemed acceptable, considering the enhanced robustness defense capability that LRR offers.

\subsection{Visualization Insights}
\label{app:extra-visualizations}

Given the template of the object of interest and an incoming frame, a tracker (e.g., SiamRPN++) aims to predict the object's position by correlating the deep features of both the template and the frame. 
An attack introduces adversarial perturbation to the frame with the intent to mislead the correlation process. 
We illustrate the comparison of LRR with and without defense visually in \Secref{subsec:ablation-and-discussion} at the image level, and we aim to delve deeper into the comparison at the correlation level.

\begin{figure}[t]
\centering
\includegraphics[width=0.925\linewidth]{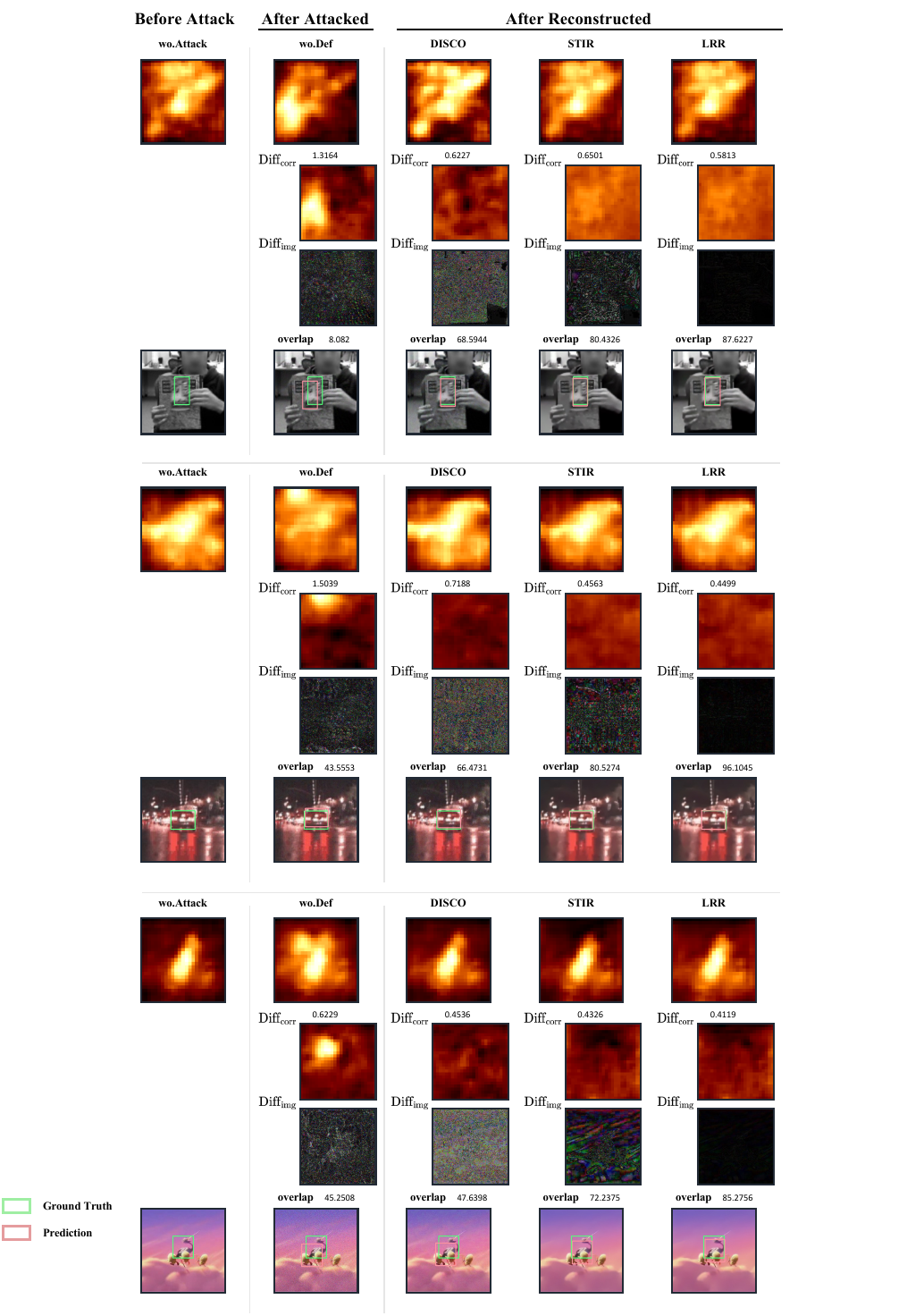}
   \caption{Visualization comparison before \& after defense from DISCO, STIR and LRR.}
\label{fig:app-vis-comparsion}
\end{figure}

More specifically, we provide visualizations in \Figref{fig:app-vis-comparsion} that demonstrate correlation maps from frames processed by our LRR method align much more closely with those unmarred by attack than other defense approaches. 
As the visualization illustrates, our LRR exhibits lower correlation map differences than DISCO \citep{ho2022disco} and STIR. 
This is because LRR effectively achieves semantic consistency between the reconstructed frame and the object template, while DISCO and STIR are primarily designed for image quality restoration and overlook the semantic consistency of the template. 
We observe that DISCO and STIR maintain relatively lower overlap with the ground truth compared to our LRR, highlighting the efficacy and precision of the LRR method in maintaining semantic integrity amidst adversarial perturbations.

\begin{figure}
\centering
\includegraphics[width=0.8\linewidth]{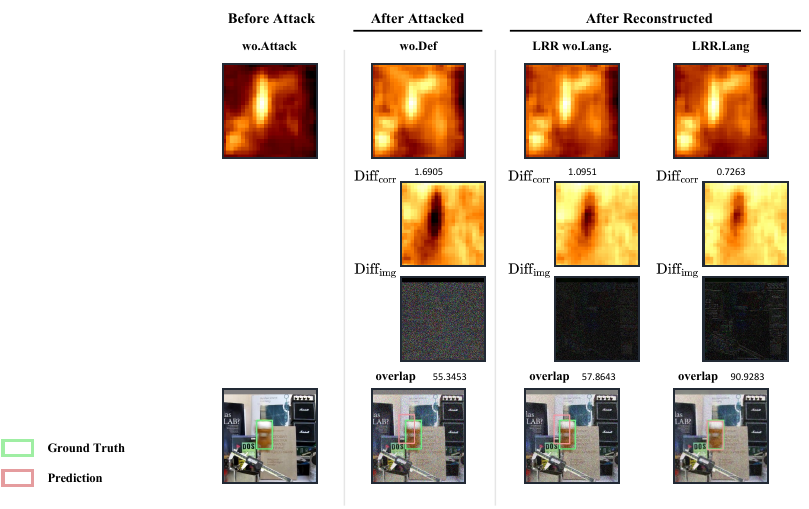}
   \caption{Visualization comparison ResampleNet with \& without language guidance when the input frame contains the object of interest.}
\label{fig:app-vis-with-target}
\end{figure}

Furthering our exploration, we investigate the impact of the resampling module in STIR, informed by the guidance of text embedding through visualization.
the LResampling module does not change the features of input frames directly but changes the rendering process of STIR. 
To elaborate, STIR can reconstruct the colors of any given coordinates as \eqref{eq:stir_sim}. Naively, we feed STIR with grid coordinates (discrete integral coordinates). 
In this context, we employ the LResampleNet to predict coordinate offsets (non-integral values) around the grid coordinates as illustrated in \eqref{eq:lresample}, and subsequently, STIR can render an image based on the predicted offsets and grid coordinates. 
LResampleNet carries dual advantages:

\textbf{If the input frame contains the object of interest,} the predicted coordinates are based on the text embedding and can highlight the object automatically. 
As the visualization of correlation maps shown in \Figref{fig:app-vis-with-target}, our final version could suppress noises from adversarial perturbations. 
Without language guidance, the overlap of $\text{LRR\ wo.Lang}$'s results with respect to the ground truth is much smaller than the $\text{LRR\ .Lang}$ ($57.86$ vs. $90.93$). 

\begin{figure}
\centering
\includegraphics[width=0.8\linewidth]{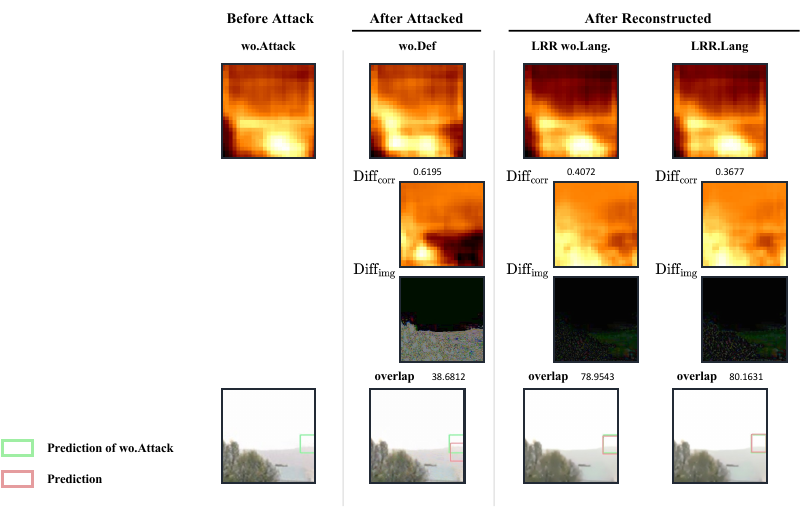}
   \caption{Visualization comparison ResampleNet with \& without language guidance when the input frame does not contain the object of interest.}
\label{fig:app-vis-without-target}
\end{figure}

\textbf{If the input frame does not contain the object of interest,} the predicted coordinates are around the grid coordinates and will keep the high restoration quality. 
As shown in \Figref{fig:app-vis-without-target}, LRR could recover the quality and make the prediction similar to the frame without attack. 

\begin{table}
\centering
\setlength{\extrarowheight}{0pt}
\addtolength{\extrarowheight}{\aboverulesep}
\addtolength{\extrarowheight}{\belowrulesep}
\setlength{\aboverulesep}{0pt}
\setlength{\belowrulesep}{0pt}
\caption{Comparing DISCO, STIR, LRR wo.Lang. and LRR under Four Attacks.}
\label{tab:app-comparsion}
\resizebox{\linewidth}{!}{%
\begin{tabular}{c|c|ccccc|ccccc|ccccc} 
\toprule
\multirow{2}{*}{SiamRPN++} & \multirow{2}{*}{Defends} & \multicolumn{5}{c|}{OTB100 Prec. (\%)} & \multicolumn{5}{c|}{VOT2019 EAO} & \multicolumn{5}{c}{UAV123 Prec. (\%)} \\
 &  & Cln. & RTAA & IoU & CSA & SPARK & Cln. & RTAA & IoU & CSA & SPARK & Cln. & RTAA & IoU & CSA & SPARK \\ 
\hline
\multicolumn{1}{l|}{} & \multicolumn{1}{l|}{} & \multicolumn{1}{l}{} & \multicolumn{1}{l}{} & \multicolumn{1}{l}{} & \multicolumn{1}{l}{} & \multicolumn{1}{l|}{} & \multicolumn{1}{l}{} & \multicolumn{1}{l}{} & \multicolumn{1}{l}{} & \multicolumn{1}{l}{} & \multicolumn{1}{l|}{} & \multicolumn{1}{l}{} & \multicolumn{1}{l}{} & \multicolumn{1}{l}{} & \multicolumn{1}{l}{} & \multicolumn{1}{l}{} \\
\multirow{5}{*}{Res50} & wo.Def & \textbf{91.4} & 32.7 & 75.9 & 47.2 & 69.8 & \textbf{0.277} & 0.080 & 0.153 & 0.089 & 0.079 & 79.5 & 41.2 & 70.5 & 46.5 & 40.8 \\
 & DISCO & 86.0 & 86.3 & 78.6 & 83.6 & 85.7 & 0.249 & 0.244 & 0.190 & 0.204 & 0.248 & 79.1 & 76.8 & 75.9 & 77.7 & 76.0 \\
 & STIR & 88.1 & 85.9 & 84.0 & 85.9 & 87.7 & 0.268 & 0.247 & 0.213 & 0.219 & 0.256 & 79.5 & 77.0 & 78.2 & 79.7 & 77.9 \\
 & LRR wo.Lang & 88.1 & 86.1 & 84.4 & 86.0 & 87.8 & 0.270 & 0.250 & \textbf{0.219} & 0.222 & 0.254 & \textbf{79.6} & 77.4 & 78.4 & 79.8 & 78.1 \\
 & LRR & 87.8 & \textbf{86.9} & \textbf{85.3} & \textbf{89.4} & \textbf{89.3} & 0.262 & \textbf{0.255} & 0.217 & \textbf{0.237} & \textbf{0.269} & 79.3 & \textbf{77.7} & \textbf{78.6} & \textbf{81.8} & \textbf{79.3} \\ 
\hline
\multirow{5}{*}{MobileV2} & wo.Def & 85.5 & 25.6 & 67.8 & 40.9 & 32.2 & \textbf{0.267} & 0.062 & 0.125 & 0.083 & 0.037 & 80.3 & 39.3 & 66.2 & 42.2 & 22.5 \\
 & DISCO & 82.9 & 78.7 & 75.0 & 79.9 & 80.1 & 0.175 & 0.161 & 0.132 & 0.166 & 0.208 & 74.6 & 76.2 & 73.3 & 72.9 & 75.3 \\
 & STIR & 85.5 & 80.7 & 79.0 & 79.8 & 83.7 & 0.238 & 0.204 & 0.163 & 0.216 & 0.225 & 78.4 & 77.7 & 74.9 & 75.7 & 78.2 \\
 & LRR wo.Lang & \textbf{86.1} & 81.2 & 79.0 & 81.1 & 84.5 & 0.238 & 0.201 & 0.151 & 0.219 & 0.228 & \textbf{79.6} & 78.0 & 75.3 & 76.0 & 78.4 \\
 & LRR & 85.6 & \textbf{83.2} & \textbf{82.1} & \textbf{83.9} & \textbf{85.4} & 0.240 & \textbf{0.205} & \textbf{0.166} & \textbf{0.223} & \textbf{0.239} & 79.1 & \textbf{78.7} & \textbf{75.9} & \textbf{76.2} & \textbf{79.0} \\
\bottomrule
\end{tabular}
}
\end{table}

Moreover, in \tableref{tab:app-comparsion}, with our LResampleNet, two variants of LRR outperform STIR and DISCO on all datasets and attacks. 

\begin{figure}
\centering
\includegraphics[width=1.0\linewidth]{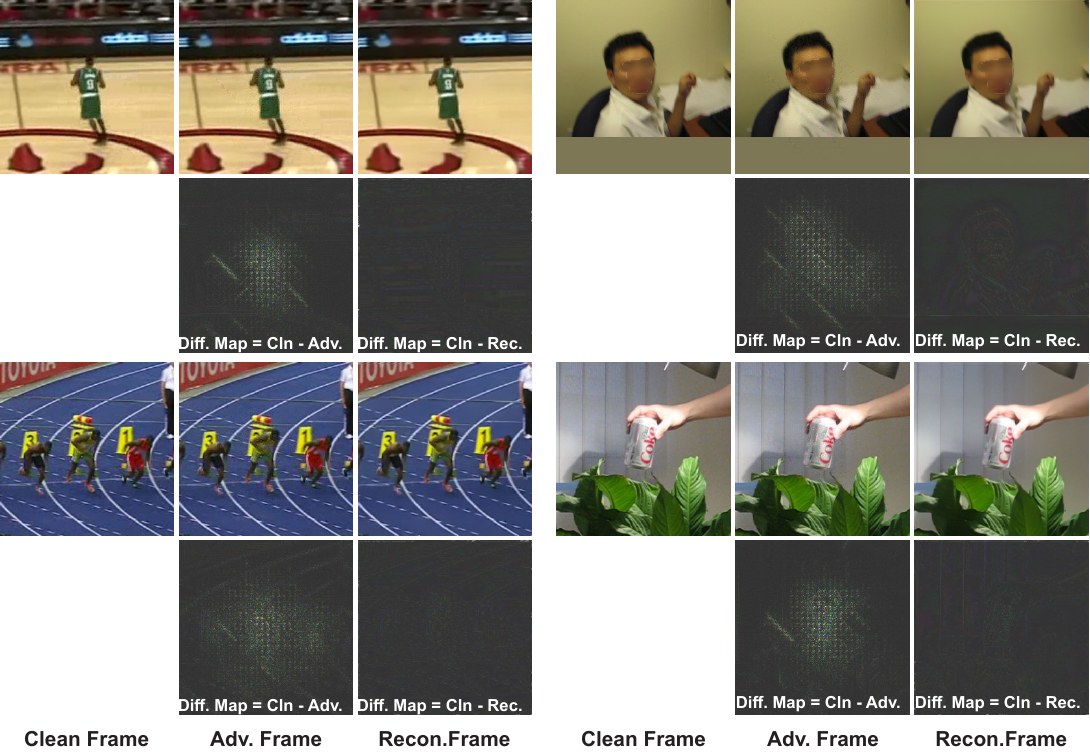}
   \caption{Visualization of clean frames, adversarial frames from CSA, and reconstructed adversarial frames based on our method. We also show the difference map between the adversarial frame and the corresponding clean frame and the difference map between the reconstructed adversarial frame and the corresponding clean frame}
\label{fig:vis-CSA}
\end{figure}

\begin{figure}
\centering
\includegraphics[width=1.0\linewidth]{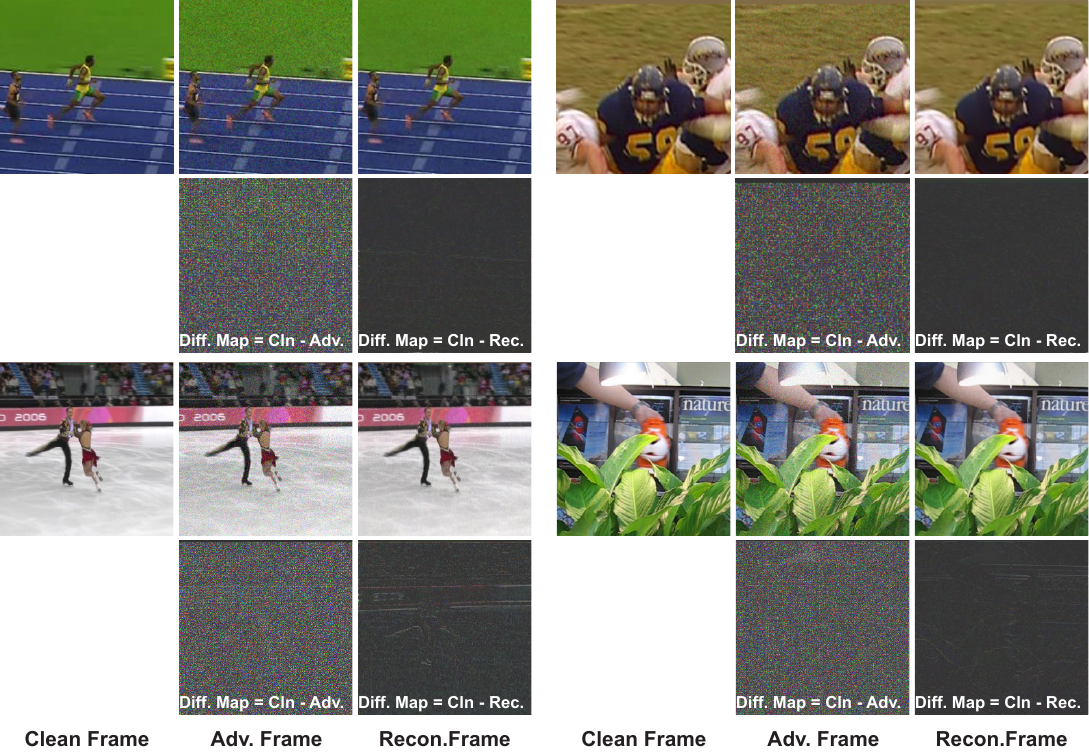}
   \caption{Visualization of clean frames, adversarial frames from IoUAttack, and reconstructed adversarial frames based on our method. We also show the difference map between the adversarial frame and the corresponding clean frame and the difference map between the reconstructed adversarial frame and the corresponding clean frame}
\label{fig:vis-IoU}
\end{figure}

\begin{figure}
\centering
\includegraphics[width=1.0\linewidth]{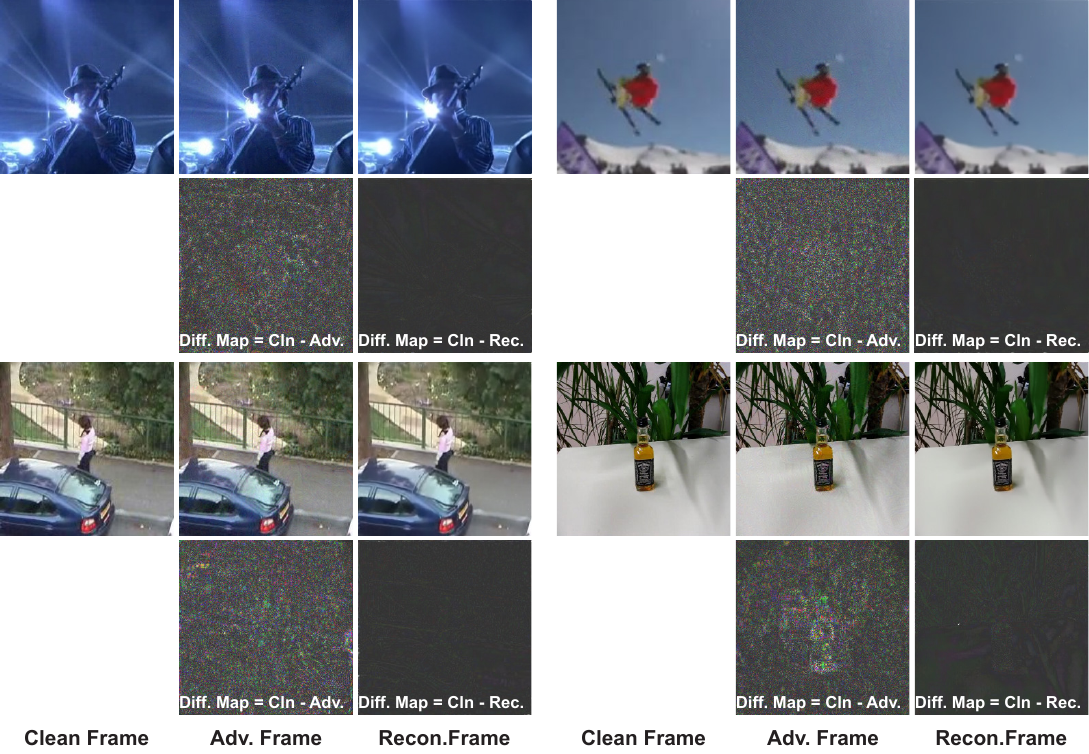}
   \caption{Visualization of clean frames, adversarial frames from SPARK, and reconstructed adversarial frames based on our method. We also show the difference map between the adversarial frame and the corresponding clean frame and the difference map between the reconstructed adversarial frame and the corresponding clean frame}
\label{fig:vis-SPARK}
\end{figure}

Furthermore, we provide the visualizations of clean frames, adversarial frames after attacks, and reconstructed frames after defense, validating the effectiveness of our method in terms of image quality variations. In particular, we consider three typical attacks, \ie, CSA, IoUAttack, and SPARK, respectively, and show their results in \Figref{fig:vis-CSA}, \Figref{fig:vis-IoU}, and \Figref{fig:vis-SPARK}. From the visualization results, we observe that: \textit{First}, The three attacks can generate adversarial perturbations with different textures according to the difference maps shown in the figures. \textit{Second}, for all attacks, our method can eliminate all adversarial perturbations effectively, though they have different textures.

\subsection{Transferability to Transformer-based Trackers}
\label{app:transformer-based-tracker}

To demonstrate the transferability of our LRR approach, we modified it for the recently proposed ToMP-50 transformer-based tracker model \citep{mayer2022transforming}, using RTAA to attack and applying LRR for defense, and assessed the outcomes across three different datasets.
As substantiated in \tableref{tab:appendix_tomp}, several observations are apparent: firstly, the application of RTAA notably degrades the accuracy of the transformer-based tracker across all three datasets. 
Secondly, despite these aggressive attacks, our method retains its robust defense capabilities, maintaining high tracking accuracy.
This illustrates the notable transferability of the LRR approach, maintaining its effectiveness even when applied to newly developed tracking models, including those based on transformer architectures.

\subsection{Comparing with Tracking Adversarial Defense}

\begin{table}[ht]
\centering
\caption{Comparison of LRR with RTAA's Defense on four datasets.}
\label{tab:appendix_rtaa_defense}
\resizebox{\linewidth}{!}{%
\begin{tabular}{c|c|ccc|ccc|ccc|ccc}
\toprule
\multirow{2}{*}{SiamRPN++} & \multirow{2}{*}{Attacks} & \multicolumn{3}{c|}{OTB100 Prec.~(\%)} & \multicolumn{3}{c|}{VOT2019 EAO} & \multicolumn{3}{c|}{UAV123 Prec.~(\%)} & \multicolumn{3}{c}{NFS30 Prec.~(\%)} \\
 &  & Org. & LRR & RTAA$_\text{Def}$ & Org. & LRR & RTAA$_\text{Def}$ & Org. & LRR & RTAA$_\text{Def}$ & Org. & LRR & RTAA$_\text{Def}$ \\ 
\hline
\multicolumn{1}{l|}{} & \multicolumn{1}{l|}{} & \multicolumn{1}{l}{} & \multicolumn{1}{l}{} & \multicolumn{1}{l|}{} & \multicolumn{1}{l}{} & \multicolumn{1}{l}{} & \multicolumn{1}{l|}{} & \multicolumn{1}{l}{} & \multicolumn{1}{l}{} & \multicolumn{1}{l|}{} & \multicolumn{1}{l}{} & \multicolumn{1}{l}{} & \multicolumn{1}{l}{} \\
\multirow{5}{*}{Res50} & wo.Atk & 91.4 & \textbf{87.8} & 72.0 & 0.277 & \textbf{0.262} & 0.197 & 79.5 & \textbf{79.3} & 65.2 & 59.9 & \textbf{56.0} & 46.6 \\
 & RTAA & 32.7 & \textbf{86.9} & 76.5 & 0.080 & \textbf{0.255} & 0.155 & 41.2 & \textbf{77.7} & 71.8 & 24.4 & \textbf{56.5} & 40.5 \\
 & IoUAttack & 75.9 & \textbf{85.3} & 56.1 & 0.153 & \textbf{0.217} & 0.136 & 70.5 & \textbf{78.6} & 57.0 & 42.0 & \textbf{55.5} & 27.0 \\
 & CSA & 47.2 & \textbf{89.4} & 31.6 & 0.089 & \textbf{0.237} & 0.079 & 46.5 & \textbf{81.8} & 41.4 & 19.6 & \textbf{58.0} & 13.1 \\
 & SPARK & 69.8 & \textbf{89.3} & 60.6 & 0.079 & \textbf{0.269} & 0.078 & 40.8 & \textbf{79.3} & 47.2 & 40.5 & \textbf{59.3} & 35.9 \\
\bottomrule
\end{tabular}
}
\end{table}

\begin{table}[ht]
\centering
\caption{Comparison of LRR with RTAA's Defense cost on OTB100.}
\label{tab:appendix_rtaa_timecost}
\resizebox{0.38\linewidth}{!}{%
\begin{tabular}{c|c} 
\toprule
Module & Defense cost per frame (ms) \\ 
\hline
 &  \\
LRR & 39 \\
RTAA$_\text{Def}$ & 215 \\
\bottomrule
\end{tabular}
}
\end{table}

We have already demonstrated the effectiveness of our proposed method compared to various defense strategies; however, exploration of other defense approaches specifically designed for visual object tracking tasks remains pending.
In this section, comprehensive comparisons with RTAA \citep{jia2020robust} are included across four datasets and against four attack methods, as illustrated in \tableref{tab:appendix_rtaa_defense}.
The RTAA defense method was implemented utilizing the codes from the official repository to defend against the aforementioned attacks.

Clearly, our method presents several notable advantages over RTAA's defense strategy.
Firstly, our method consistently achieves superior tracking accuracy compared to RTAA's defense method against all types of attacks and across all datasets examined.
Secondly, the impact of our method on clean data is minimal, preserving the integrity and accuracy of the unaffected data. 
In contrast, RTAA's defense method could notably diminish accuracy when applied to clean data.
Additionally, a comparative analysis of the time costs between LRR and RTAA on OTB100 is provided in \tableref{tab:appendix_rtaa_timecost}.
This comparison elucidates the enhanced efficiency of our method over RTAA, strengthening the argument for its application in practical, time-sensitive scenarios. 
The methodical implementation and rigorous evaluation underscore the robustness and reliability of our method, validating its potential as a superior defense mechanism in visual object-tracking tasks.

\subsection{Detailed Discussion of Defense Efficiency}
\label{app:defense-efficiency}

In \Secref{subsec:ablation-and-discussion}, we report both the time costs of our methods and the attack costs of the attackers in \tableref{tab:time_cost}, respectively.
We demonstrate that our proposed methods exhibit superior frame processing efficiency compared to most attackers, with the exception of CSA \citep{yan2020cooling}, which employs a fast perturbation generator.
Furthermore, our LRR surpasses STIR in adversarial attack defense capability, sacrificing only a negligible amount of efficiency—4ms per input frame defense. 
In the case of less efficient attackers such as IoU Attack \citep{jia2021iou} and RTAA \citep{jia2020robust}, we receive attacked frame sequences at a rate of less than 0.1 frames per second (fps). 
In this context, the computational cost added by LRR is practically negligible. 
For more efficient attackers, such as SPARK \citep{guo2020spark} and CSA, under the assumption that the attacker and defender utilize the same computational resources, our LRR method trades off a portion of tracking efficiency in favor of a significant improvement in the tracker's robustness.
In real-world scenarios, where attackers and defenders are typically deployed on separate systems, our STIR defense sustains online frame processing at an approximate rate of 29 fps, while LRR functions at around 25 fps.

Moreover, computation time costs can be further optimized by adapting the defense policy. For instance, by employing the target overlap ratio compared to the previous frame as a threshold, we can bypass processing for 25\% of frames and still maintain an overlap ratio not lower than 85\%.

\subsection{Feasibility of using diffusion for tracking defense} 
\label{app:diffpure}
We explore the efficacy of the recently developed diffusion-based adversarial defense method, DiffPure \citep{nie2022diffusion}, for tracking defense. Specifically, we apply DiffPure to safeguard against three attacks, i.e., RTAA, IoUAttack, and CSA, that are used to attack the SiamRPN++ Res50 tracker on the OTB100 dataset. In our empirical study, we use DiffPure's default parameters for defense but vary the number of iterative time steps (i.e., T=1, 10, 50).

\tableref{tab:appendix_diffpure} illustrates that the three DiffPure variants enhance the precision of the tracker under different attacks, albeit to a lesser extent compared to our approach, LRR. Notably, DiffPure(T=50) is 86.9 times slower than LRR, requiring an average of 3391 ms for each frame, rendering it nearly impractical for tracking tasks. Even with a reduced time step to 1, DiffPure speeds up to 146 ms per frame, still 3.7 times slower than LRR. It is essential to note that the default DiffPure configuration sets T=100 time steps for purification, which is impractical for tracking tasks due to time constraints. In conclusion, further investigation is needed to understand the potential of leveraging diffusion for tracking defense.
\begin{table}[h]
\centering
\vspace{-10pt}
\caption{Comparing DiffPure \cite{nie2022diffusion} with LRR on OTB100 where we use them to defend CSA and RTAA for the SiamRPN++ Res50.}
\label{tab:appendix_diffpure}
\resizebox{0.6\linewidth}{!}{%
\begin{tabular}{c|cccc|c} 
\toprule
Defense method & Cln. & RTAA & IoUAttack & CSA & Time (ms) \\ 
\hline
\multicolumn{1}{l|}{} & \multicolumn{1}{l}{} & \multicolumn{1}{l}{} & \multicolumn{1}{l}{} & \multicolumn{1}{l|}{} & \multicolumn{1}{l}{} \\
w.o. Defense & 91.4 & 32.7 & 75.9 & 47.2 & - \\
LRR & 87.8 & \textbf{86.9} & \textbf{85.3} & \textbf{89.4} & 39 \\
DiffPure(T=1) & 87.9 & 52.3 & 78.5 & 83.5 & 146 \\
DiffPure(T=10) & 88.1 & 53.7 & 78.8 & 84.1 & 742 \\ 
DiffPure(T=50) & 88.2 & 54.2 & 79.0 & 84.3 & 3391 \\
\bottomrule
\end{tabular}
}
\end{table}

\subsection{Comparing with resizing and compression-based Defenses}
\label{app:baseline_resize_compress}

We implemented a resizing-based defense using the `cv.resize' operation in OpenCV. Specifically, for an input image $\mathbf{I}\in R^{H\times W}$, we first downsample it by a factor $r$ and get image $\mathbf{I}_\downarrow\in R^{rH\times rW}$. Then, we upsample it to the raw resolution, generating the reconstructed image $\hat{\mathbf{I}} \in R^{H\times W}$. Following this, we input the reconstructed images into trackers.

To assess the effectiveness of resizing-based defense, we varied the downsampling ratio within the range $r\in \{0.9, 0.8,\ldots,0.1\}$. As shown in \tableref{tab:appendix_resizing}, we observe that:
1. Resizing proves to enhance the tracker's accuracy under various attacks.
2. The efficacy of this enhancement varies depending on the attack type. Resizing significantly mitigates the impact of the SPARK attack, elevating precision from 69.8 to 83.9, but exhibits limited effectiveness against the RTAA, where precision increases modestly from 32.7 to 49.
3. The influence on RTAA remains constrained as precision increases from 32.7 to 49.3. Gradually increasing $r$ improves precision under RTAA but adversely affects precision in clean data and IoUAttack scenarios.
4. Compared to the resizing method, LRR consistently improves tracker precision across all attacks, showcasing a noteworthy advantage while maintaining a high score on clean data.

\begin{table}
\centering
\caption{Comparison of resizing-based defense with different settings of $r$ on OTB100.}
\label{tab:appendix_resizing}
\resizebox{\linewidth}{!}{%
\begin{tabular}{c|c|ccccccccccc} 
\toprule
\multirow{2}{*}{SiamRPN++} & \multirow{2}{*}{Attacks} & \multicolumn{11}{c}{OTB100 Prec.~(\%)} \\
 &  & Org. & LRR & $r=0.9$ & $r=0.8$ & $r=0.7$ & $r=0.6$ & $r=0.5$ & $r=0.4$ & $r=0.3$ & $r=0.2$ & $r=0.1$ \\ 
\hline
\multicolumn{1}{l|}{} & \multicolumn{1}{l|}{} & \multicolumn{1}{l}{} &  & \multicolumn{1}{l}{} & \multicolumn{1}{l}{} & \multicolumn{1}{l}{} & \multicolumn{1}{l}{} & \multicolumn{1}{l}{} &  &  &  &  \\
\multirow{5}{*}{Res50} & wo.Atk & 91.4 & 87.8 & 86.5 & 86.2 & 85.9 & 85.4 & 85.3 & 82.7 & 82.7 & 80.8 & 69.9 \\
 & RTAA & 32.7 & 86.9 & 49.3 & 56.7 & 62.0 & 69.0 & 72.5 & 77.3 & 80.2 & 80.3 & 69.0 \\
 & IoUAttack & 75.9 & 85.3 & 80.3 & 80.1 & 79.0 & 79.0 & 76.1 & 76.5 & 74.9 & 72.0 & 63.3 \\
 & CSA & 47.2 & 89.4 & 71.6 & 80.3 & 84.6 & 86.0 & 83.8 & 83.1 & 83.0 & 81.8 & 69.1 \\
 & SPARK & 69.8 & 89.3 & 83.9 & 85.1 & 88.1 & 87.8 & 86.0 & 87.7 & 85.4 & 82.5 & 72.2 \\
\bottomrule
\end{tabular}
}
\end{table}

Regarding compression, we utilize JPG compression for image reconstruction, adjusting compression qualities with
$q\in[98\%,96\%,94\%,92\%,90\%]$. The results are presented in \tableref{tab:appendix_compression}, and the following observations are made:
1. Compression with a high-quality requirement exhibits limited influence on various attacks.
2. As the compression quality decreases, precision on different attacks increases, highlighting the effectiveness of compression as a defense mechanism against adversarial tracking.
3. The improvements achieved by compression under attacks are limited and fall short of the results obtained with LRR.

\begin{table}
\centering
\caption{Comparison of compression-based defense with different settings of $q$ on OTB100.}
\label{tab:appendix_compression}
\resizebox{0.8\linewidth}{!}{%
\begin{tabular}{c|c|ccccccc} 
\toprule
\multirow{2}{*}{SiamRPN++} & \multirow{2}{*}{Attacks} & \multicolumn{7}{c}{OTB100 Prec.~(\%)} \\
 &  & Org. & LRR & $q=98\%$ & $q=96\%$ & $q=94\%$ & $q=92\%$ & $q=90\%$ \\ 
\hline
\multicolumn{1}{l|}{} & \multicolumn{1}{l|}{} & \multicolumn{1}{l}{} &  & \multicolumn{1}{l}{} & \multicolumn{1}{l}{} & \multicolumn{1}{l}{} & \multicolumn{1}{l}{} & \multicolumn{1}{l}{} \\
\multirow{5}{*}{Res50} & wo.Atk & 91.4 & 87.8 & 90.8 & 89.6 & 90.2 & 89.7 & 90.1 \\
 & RTAA & 32.7 & 86.9 & 33.5 & 42.9 & 50.1 & 60.6 & 66.1 \\
 & IoUAttack & 75.9 & 85.3 & 74.8 & 74.4 & 76.5 & 76.2 & 77.1 \\
 & CSA & 47.2 & 89.4 & 49.0 & 51.4 & 51.8 & 56.3 & 58.7 \\
 & SPARK & 69.8 & 89.3 & 78.1 & 82.1 & 83.6 & 85.4 & 85.9 \\
\bottomrule
\end{tabular}
}
\end{table}

\subsection{Details of Adversarial Tracking Attacks}
\label{app:trackingattacks}

We implement adversarial tracking attacks via the released codes from existing tracking adversarial attacks (\ie, RTAA \citep{jia2020robust}, IoUAttack \citep{jia2021iou}, CSA \cite{yan2020cooling}, and SPARK \cite{guo2020spark}) to implement attacks in our experiments. We detail some setups as follows.

For RTAA, we utilized their originally released code (\url{https://github.com/VISION-SJTU/RTAA/blob/main/DaSiamRPN/code/run_attack.py}). The process follows these steps: 1. RTAA receives an incoming image and the target location where the image is the search region cropped by the studied tracker. 2. RTAA adds adversarial perturbations to the search region and outputs an adversarial example for the tracker to handle. At each frame, the attack optimizes the adversarial perturbation iteratively ten times, with the maximum perturbation set to 10/255. 3. RTAA outputs the optimized adversarial example as the new search region.

For the IoU Attack, we adhered to their default setups in their released code for conducting our experiments (\url{https://github.com/VISION-SJTU/IoUattack/blob/main/pysot/tools/test_IoU_attack.py}). Specifically, we follow the subsequent steps: 1. IoUAttack receives the frame and targeted bounding box as inputs. 2. IoUAttack optimizes the perturbations iteratively until the IoU score is below the predefined score (See the released code for details). 3. IoU outputs the optimized adversarial frame to attack the tracker.

For CSA, we employed their released pre-trained perturbation generator to attack each frame (\url{https://github.com/MasterBin-IIAU/CSA/blob/efd69a5705dd21c6701fd4ae7922f3a44647069a/pysot/pysot/tracker/siamrpn_tracker.py}). Specifically, CSA receives the clean search region and feeds it to the pre-trained perturbation generator. Then, the generator outputs the adversarial perturbation added to the clean search region.

In the case of SPARK (\url{https://github.com/tsingqguo/AttackTracker/blob/main/tools/attack_oim.py}), we employed the targeted attack approach provided in SPARK's default setup from their released code for attacks. The procedure involves the following steps: 1. SPARK takes the search region, cropped from the input frame, the targeted trajectory, and the targeted tracker as inputs. 2. SPARK optimizes the perturbations, iterating 10 times every 30 frames and 2 times at other frames. The maximum perturbation allowed is 0.3. 3. SPARK generates the optimized adversarial search region to attack the tracker.

\end{document}